\pdfoutput=1

\documentclass[11pt]{article}

\usepackage[]{EACL2023}
\usepackage{xcolor}
\usepackage{times}
\usepackage{latexsym}
\usepackage{float}

\usepackage[T1]{fontenc}
\usepackage{booktabs}
\usepackage{csvsimple}

\usepackage[utf8]{inputenc}

\usepackage{microtype}

\usepackage{inconsolata}

\usepackage{tabularray}
\usepackage{tabularx}
\usepackage{ragged2e}

\usepackage{amsmath}

\usepackage{graphicx}

\usepackage{subfigure}

\usepackage{algpseudocode}
\usepackage{algorithm}
\usepackage{todonotes}

\title{Anchor Points: Benchmarking Models with Much Fewer Examples}

\author{Rajan Vivek, Kawin Ethayarajh, Diyi Yang, Douwe Kiela \\ Stanford University \\
\{rvivek, kawin, diyiy, dkiela\}@stanford.edu}

\begin{document}
\maketitle
\begin{abstract}

Modern language models often exhibit powerful but brittle behavior, leading to the development of larger and more diverse benchmarks to reliably assess their behavior. Here, we suggest that model performance can be benchmarked and elucidated with much smaller evaluation sets. We first show that in six popular language classification benchmarks, model confidence in the correct class on many pairs of points is strongly correlated across models. We build upon this phenomenon to propose \emph{Anchor Point Selection}, a technique to select small subsets of datasets that capture model behavior across the entire dataset. Anchor points reliably rank models: across 87 diverse language model-prompt pairs, evaluating models using 1-30 anchor points outperforms uniform sampling and other baselines at accurately ranking models. Moreover, just a dozen anchor points can be used to estimate model per-class predictions on all other points in a dataset with low error, sufficient for gauging where the model is likely to fail. Lastly, we present \emph{Anchor Point Maps} for visualizing these insights and facilitating comparisons of the performance of different models on various regions within the dataset distribution. 

\end{abstract}

\section{Introduction}

Language models have unlocked incredible generalization through scaling up parameters and pretraining data. Yet these same systems prove to be brittle, spurring the development of larger, more diverse, and more shrewd benchmarks to reliably assess their behavior \cite{qnli, mmlu, dynabench, recogs}. Modern benchmarks typically have on the order of $10^5$ validation examples, with $10^3 - 10^4$ per task. Such numbers ensure that validation performance strongly correlates with test performance and that all in-domain regions are captured. But these sizes are often unwieldy for rapid experimentation and do not easily afford interpretability. To compare model configurations, design the most robust prompt, or analyze failure cases, researchers and practitioners must forward-pass (and potentially manually inspect) the development set many times. How small can benchmark development sets be while still capturing model behavior over the full breadth of the benchmark? Surprisingly, very small---just several to a few dozen examples might suffice.

\begin{figure}
  \centering
    {\includegraphics[scale=0.4]{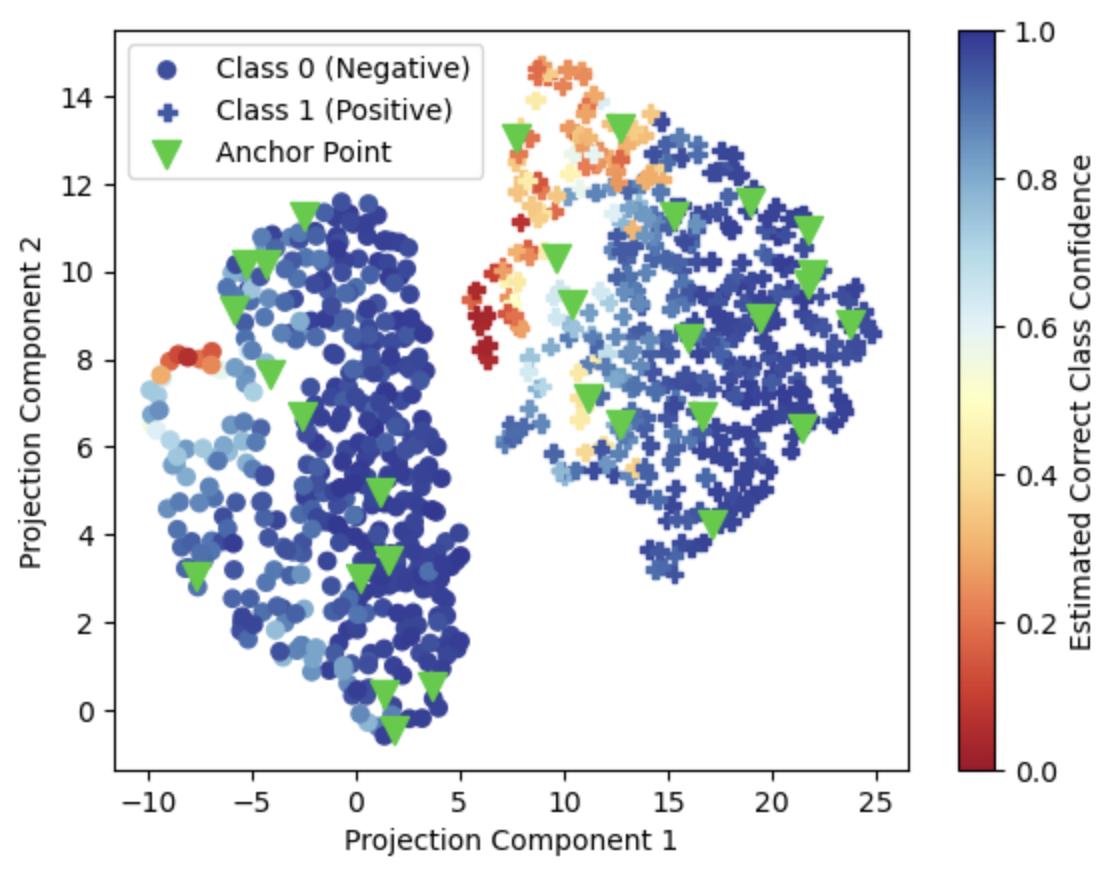}} 
  \caption{SST-2 Validation Set Anchor Point Map. The locations of all 872 points are learned using the predictions of $N=60$ randomly-selected source models on SST-2. (Note $N$ can be as small as 10; see Table \ref{tab:agreement_table}). We then evaluate a held-out model, Falcon-7B, on 30 anchor points (green triangles). The model's predictions on only these 30 points are used to estimate the Falcon-7B predictions on the remaining 842 points with a mean absolute error of 0.09, achieving 92\% agreement with the model's true predictions. The anchor points identify regions where the model is weak (red regions). We show the same Anchor Point Map colored by the true Falcon-7B predictions in Figure \ref{fig:sst2-corrmap-comparison}, demonstrating that the model is indeed weak in these areas.}
  \label{fig:main-corrmap}
  
\end{figure}

In this work, we investigate the problem of benchmarking model performance and revealing model weaknesses on large datasets with as few evaluation examples as possible, an objective we call \emph{micro-benchmarking}. We propose \emph{Anchor Point Selection}, a technique that finds small evaluation sets that are maximally representative of model behavior over the entirety of a large dataset. We find that anchor points are effective development sets: across 87 diverse language models and prompts, using 1-30 anchor points shows superior performance at ranking model performance over random and embedding-based selection baselines. Moreover, evaluating just a dozen anchor points can be used to predict the model's instance-level predictions on all other points in the dataset with high agreement on average, sufficient to estimate where models are likely to fail. 

\begin{figure}[t]
\centering
\includegraphics[width=7cm]{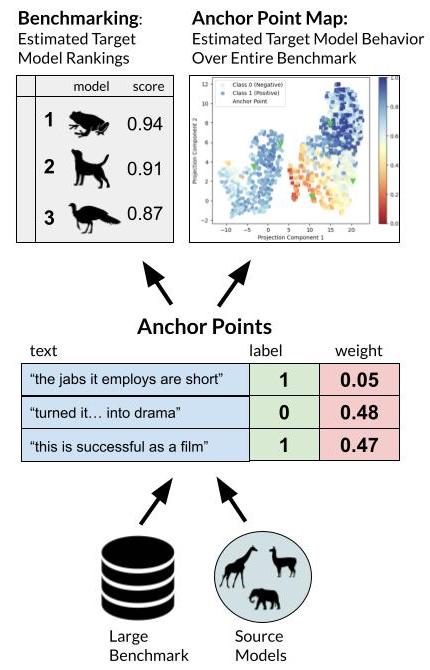}
\caption{Anchor Points are Micro-Benchmarks, tiny representative subsets of large benchmarks. Correlative structure in the predictions of source models on the large benchmark can be used to extract these points. Each anchor point has a weight corresponding to the fraction of the benchmark it represents. Evaluating models on the anchor points produces a score that rank correlates with performance on the entire benchmark. Anchor Point Maps visualize a given model's likely instance-level performance on all points in the benchmark using only its performance on the anchor points. \label{fig:main}}

\label{fig:test}
\centering
\end{figure}

Our approach builds upon a simple insight: for many pairs of points from a given dataset, the predicted probability of the correct class strongly correlates across models. Thus, evaluating every model on the entirety of a dataset is redundant. The predictions of existing \emph{source} models on a given dataset are telling about predictions of a new \emph{target} model, even across considerable performance gaps. This mirrors phenomena like \emph{Accuracy on the Line}~\cite{accline} and \emph{Agreement on the Line}~\cite{agreement}, but at a much more granular scale. Anchor Points can be selected by identifying dataset medoids in a space where distance is a positive, monotonically decreasing function of cross-model correlations between points. We further propose to visualize this space using multi-dimensional scaling  and U-MAP \cite{umap}, which we call \emph{Anchor Point Maps}. Anchor Point Maps show what region each anchor point captures and highlight that different models and prompts may struggle on different dataset regions, allowing fine-grained comparisons.

Emphatically, we do not aim to replace large benchmarks, but rather provide cheaper signal about model performance on these benchmarks to help accelerate the development of models and prompts. We also strive to be blunt about the limitations of our technique. Anchor point generalization relies on the predictive correlations of source models being consistent with that of target models. This is not always the case, leading to poor generalization to specific target models depending on the choice of source models. We share simple rules of thumb to promote this source-target consistency, but lack rigorous theory to guide source model selection.  We lay foundations for further research into model predictive correlations and efficient, interpretable model evaluation.

\section{Related Work}

\paragraph{Sample-Efficient Model Evaluation}

We deviate from prior sample-efficient model evaluation literature along key axes. Many works~\cite{active-surrogate, active-testing, nolabels, withouttest} minimize evaluation \emph{annotation} costs by actively selecting points to annotate for evaluating a given model. In the era of large benchmarks and large models, labeled evaluation examples are widely-available but evaluating all of them is cumbersome. Our technique instead minimizes the number of \emph{forward-passed examples} necessary for reliable model evaluation. Furthermore, our selection strategy is agnostic to the target model(s) being evaluated. The resulting evaluation set is transferable to other language models.

Other works \cite{pedro, kawin, fix-benchmarking} point out that it is often a minority of points that differentiate the performance of various models. \citet{pedro} show that evaluating models on points identified as the most \emph{discriminative} can effectively rank model performance. However, optimizing for discriminability does not result in representative subsets: very easy or very hard points will tend to be excluded, leading to a different distribution. Instead, we optimize for \emph{representativeness} which we show naturally leads to discriminability.

\paragraph{Instance-Level Model Performance}

Despite being noisy \cite{instance-level}, instance-level model predictions are a rich source of information about model behavior. \citet{cartography} present Data Maps, a powerful technique that leverages instance-level predictions to reveal underlying structure in the interplay of models and data points. Various training example regions play distinct roles in guiding a classifier to its solution. Unlike our technique, Data Maps are not used for comparing model performance or isolating distinct regions of the dataset distribution where models are weak. 

\citet{kawin} further show that instance-level predictions can be used to quantify how much information a given model can extract from a dataset, providing a formal metric of dataset difficulty that exposes model behavior at the dataset and data point level.

\paragraph{Predictive Correlations across Models}

\citet{accline} present the \emph{Accuracy on the Line} phenomenon: out-of-distribution (OOD) performance is strongly positively correlated with in-distribution (ID) performance for a wide range of models and distribution shifts. This unexplained by classical theory which provides only weak bounds relating the two metrics \cite{id-ood}. \citet{agreement} further show that the \emph{agreement} of two classifiers on $ID$ data strongly linearly correlates with their agreement on $OOD$ data whenever \emph{Accuracy on the Line} holds. Both phenomena can be used to cheaply estimate model $OOD$ performance. Our findings presented in this work mirror these phenomena at a much more granular scale, i.e., data instances rather than data sets.

\paragraph{Coresets}

Anchor Points can be interpreted as a coreset, broadly defined as a small "summary" of a large set of data~\cite{feldman-intro, coresets}. Selecting dataset coresets for efficient model training is widely-explored in prior work, performed through various techniques including clustering~\cite{kmedoids,kcenter}, gradient-matching~\cite{data-efficient,gradmatch}, bi-level optimization~\cite{bilevel, glister}, and submodularity-based methods~\cite{submodular}. We use a clustering-based approach, but focus on summarizing data for model \emph{evaluation} rather than training. 

\section{Problem Set-Up}

Let $\mathcal{D} = \{(x_i, y_i)\}_{i=1}^K$ be a language classification benchmark. $\mathcal{D}$ is partitioned into a training split $\mathcal{D}_{train}$ and one or more evaluation splits $\mathcal{D}_{eval}$, each drawn i.i.d from $\mathcal{D}$. We are given $M$ models to evaluate, denoted as the target set $\mathcal{T} = \{\phi_m\}_{m=1}^M$. Each unique $\phi_m$ corresponds to a model fine-tuned on $\mathcal{D}_{train}$ or paired with a specific prompt template that directs the model at solving $\mathcal{D}$. The dataset has an evaluation metric (accuracy, F1-score, etc.) to measure the aggregate performance $P_m$ of each $\phi_m$ on an evaluation split $\mathcal{D}_{eval}$. 

We additionally have access to the instance-level predictions of a source set of $N$ model and prompt template pairs $\mathcal{S} = \{\phi_n\}_{n=1}^N$ over the entirety of $\mathcal{D}_{eval}$, where $N \geq 10$ in practice. $\mathcal{S}$ and $\mathcal{T}$ are disjoint. The predictions occupy an $N \times |\mathcal{D}_{eval}| \times Y$ tensor $P_\mathcal{S}$, where $Y$ is the cardinality of the classification task. In practice, the $N$ source models could be open-source models that can be run locally for free while target models might be closed-source and/or more expensive. Note that the one-time cost of evaluating these $N$ models is offset by the innumerable amount of development-time evaluations that can be replaced with evaluations over the representative subset we will extract.

\paragraph{Towards Micro-Benchmarking}
We aim to extract a development set from a large benchmark that captures the benchmark's broad coverage and reliable ranking power while being as small as possible, improving performance interpretability and evaluation efficiency. We refer to this objective as \textbf{micro-benchmarking}. The technique must 1) acquire a small representative subset of evaluation points from the large benchmark ($X_{acq} \subseteq \mathcal{D}_{eval}, |X_{acq}| \ll |\mathcal{D}_{eval}|$), 2) leverage the predictions of target models $\{\phi_m\}_{m=1}^M$ on the subset to produce scores $S_{1...M}$ that correlate as much as possible with model performances $P_{1...M}$ on the entire dataset $\mathcal{D}_{eval}$, and 3) estimate instance-level performance on untested points from $\mathcal{D}_{eval}.$ Note that the third objective is not essential to model benchmarking, but aids in achieving greater interpretability of model behavior.

\section{Micro-Benchmarking Approach}

We measure how well an example $(x_1,y_1)$ represents another example using the Pearson correlation of correct class confidence in the two examples measured across the predictions of the source models in $\mathcal{S} = \{\phi_n\}_{n=1}^{N}$. If our data is not annotated, we can instead compute $\mathcal{Y}$ Pearson correlations (one across each class) per example pair and average them. However, model confidences are bounded by [0, 1] and thus can only follow a linear trend for a bounded range. Akin to \citet{accline}, we take the logit transform of the confidences to scale the axes from [0,1] to $[-\inf, +\inf]$ prior to computing correlations. We denote this composite function as $CORR_\mathcal{S}$. The task of selecting a $X_{acq}$ that is maximally representative of $D_{eval}$ for all models in $\mathcal{S}$ reduces to solving a K-Medoids problem that maximizes the correlation between the selected and the remaining points: 

\begin{equation}
    \min_{X_{acq}}\sum_{i = 1}^B\sum_{x_{eval}^j \in Q_i} 1 - CORR_{\mathcal{S}}(x_{acq}^i, x_{eval}^j)
    \label{eq:objective-2}
\end{equation}

where $B$ is the number of anchor points and $Q_i$ is the set containing all $x_{eval}$ that are more strongly correlated with $x_{acq}^i$ than any other $x_{acq}^j$. This objective makes the assumption that the estimate for $\phi_m(x_{eval})$ leverages only the most strongly correlated $\phi_m(x_{acq}^i)$. We efficiently solve this objective using the Partitioning Around Medoids (PAM) algorithm \cite{kmedoids}. 

\subsection{Anchor Point Techniques}

We present two techniques for leveraging anchor points to benchmark model performance with minimal evaluation examples.

\paragraph{Technique 1: Anchor Point Predictor}

We propose the Anchor Points Predictor, an ensemble of univariate linear regression models that use model predictions on each anchor point to estimate predictions on all other points in  $X_{eval}$. Specifically, anchor point $\phi(x_{acq}^i)$ is used to estimate all $\phi(x_{eval}) \in Q_i$. These instance-level prediction estimates can then be used to detect regions of model weaknesses (see Section \ref{corr_map_intro}) and compute an estimate of any performance metric for the target model. This technique requires that the trend lines fitted to source model predictions explains the variance in target model predictions well. We explore when this is the case in Appendix \ref{appendix:generalization}. Algorithms \ref{alg:fit} and \ref{alg:predict} show pseudocode for fitting and making predictions with the Anchor Point Predictor Model.

\paragraph{Technique 2: Anchor Point Weighted Score}

Rather than attempting to estimate exact model performance at the instance-level, we can aim to produce a score for each target model that highly correlates with the model's performance on $\mathcal{D}_{eval}$. Using the predictions of the source model set $\mathcal{S}$, we select $N \in {1...B}$ anchor points according to Equation \ref{eq:objective-2}. Each of these points $x_{acq,i}$ strongly correlates with a subset of the untested points in $\mathcal{D}_{eval}$, namely all $x_{eval}^j \in Q_i$. We propose the Anchor Point Weighted (APW) score, a weighted average of the model's correct class probability predictions on the anchor points with weights proportional to cluster size.

\begin{equation}
    APW(\phi_i) = \frac{1}{|\mathcal{D}_{eval}|}\sum_{i=1}^B |Q_i| * \phi_i(x_{acq}^i)[y_i]
    \label{eq:apw}
\end{equation}

\paragraph{Anchor Point Maps}
\label{corr_map_intro}

To highlight the insights provided by anchor points, namely model strengths and weaknesses in distinct regions of a dataset, we propose to visualize the cross-model correlative space from which anchor points are drawn. We compute pairwise correlations between all points using $CORR_{\mathcal{S}}$, creating a $|\mathcal{D}_{eval}| \times |\mathcal{D}_{eval}|$ correlation matrix $C$. We then represent the distance matrix $Z$ as a positive, monotonically decreasing function of correlation: $Z = 1 - C$. Finally, we cast the points to a high-dimensional continuous space using Multi-Dimensional Scaling on $Z$ and project the space to two components with U-MAP~\cite{umap}. This technique visualizes the dataset in a space where a model's performance on each point generalizes to the local neighborhood of that point.

\begin{figure*}[t]
  \centering
  \subfigure[Pairwise Correlation Matrix of ALL-Family Correct Class Confidences on 50 QQP Examples. Rich structure indicates the relatedness of various examples. Many are strongly correlated.\label{fig:mrpc-matrix}] {\includegraphics[scale=0.35]{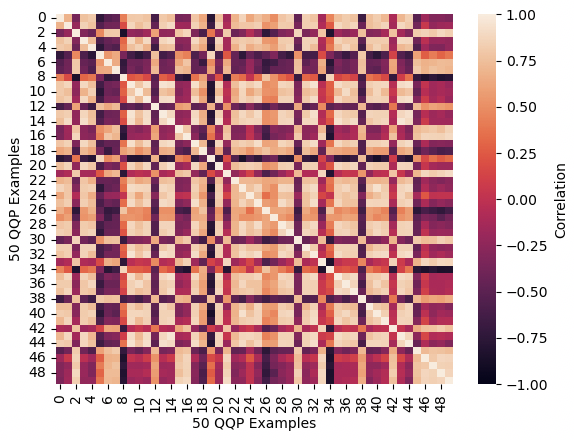}} \quad
  \subfigure[Correct Class Confidences of 87 Language Model Predictions from all model families on Two Selected QQP examples. A model correctly classifying one example is predictive of the model correctly classifying the other. This phenomenon is very common. \label{fig:QQP_pair_final}]{\includegraphics[scale=0.35]{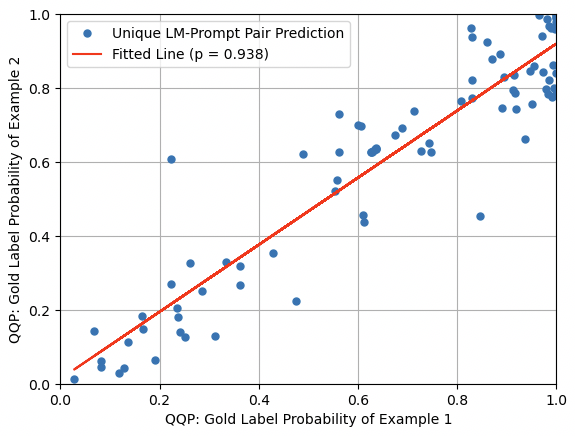}}

  \caption{Predictive Correlations at the Instance-Level Across Language Models}
  \label{fig:corrs}
  
\end{figure*}

\section{Predictive Correlations}

To motivate our approach, we first show that language models make predictions with consistent structure. For many pairs of points from a given dataset, the predicted probability of the correct class strongly correlates across models.

\paragraph{Experimental Set-Up}

We obtain a diverse set of language models from HuggingFace and the OpenAI API: 27 BERT-family models, 11 non-instruction-tuned GPT-family models, 5 instruction-tuned GPT-family models, and 5 GPT-3/3.5 variants (all listed in Tables \ref{tab:accs1} - \ref{tab:accs2}). We denote these model sets as the BERT-family, GPT-family, InstructGPT-family, and OpenAI-family respectively. For six datasets from the GLUE benchmark \cite{glue}, we finetune each BERT-family model on the training split $\mathcal{D}_{train}$ and obtain its predicted probability of the correct class on each instance in the validation split $\mathcal{D}_{eval}$. For the other model families we zero-shot prompt  for the same datasets using three multiple-choice prompts per dataset (Appendix \ref{appendix:prompts}). We obtain the predicted probability of each class by performing softmax over the log probabilities of candidate answer sequences. Thus, we obtain a prediction matrix $P$ for each model family and dataset with size $N_{subjects} \times |\mathcal{D}_{eval}|$ matrix where $N_{subjects} = 27$ for $P_{BERT}$, $30$ for $P_{GPT}$, $15$ for $P_{IGPT}$, $15$ for $P_{OAI}$, and 87 for their union $P_{ALL}$.

\paragraph{Model predictions on pairs of examples are linearly correlated across models.}

The prediction matrices are approximately low-rank for all tasks~(Table \ref{table:info_table}). This can be attributed to a simple phenomenon: let $\phi_n(x_1)[y]$ denote the probability mass (equivalently, model confidence) outputted by $\phi_n$ on $x_1 \sim \mathcal{D}_{eval}$ for a given class $y$. We see that $\phi_n(x_1)[y]$ often linearly correlates with the model's prediction on a different instance $\phi_n(x_2)[y]$ across models, i.e. $\phi_n(x_2)[y] \approx w\phi_n(x_1)[y] + b$, where $w$ and $b$ can be found by fitting a trend line through the correct class confidences of all source models.

Figure \ref{fig:corrs} visualizes this phenomenon. We first observe that the correlation matrices of model predictions on the same evaluation set show rich structure. Figure \ref{fig:QQP_pair_final} shows a QQP example pair that is strongly positively correlated across models from all families. This suggests that evaluating a model on both examples is redundant. This phenomenon is wide-spread: it holds for many pairs of examples across the six GLUE tasks and 87 widely-used language models and prompts. 

\paragraph{Predictive Correlations Tend to Generalize Across Diverse Models}

Figure \ref{fig:transfer-table} in Appendix \ref{appendix:generalization} shows how well the strong trends from each family transfer to other families. Errors are lowest in the bottom row, suggesting that using source models that span all model families result in the most generalizable trend lines. We also observe that the predictive correlations of the BERT-family do not generalize well to other model families, which we explore further in Figure \ref{fig:contentious}. However, most errors in the table are generally low, suggesting that predictive correlations tend to generalize across diverse models.

\begin{table}[t]
\begin{center}
\begin{tabular}{l|l|l|l|l|l}%
\bfseries Task & \bfseries BERT & \bfseries GPT & \bfseries IGPT & \bfseries OAI & \bfseries ALL \\ 

\hline
MNLI & $2_{0.1}$ & $1_{0.07}$& $3_{0.07}$ & $4_{0.1}$ & $3_{0.09}$ \\
SST-2 & $1_{0.08}$ & $2_{0.05}$& $2_{0.09}$ & $3_{0.1}$ & $2_{0.09}$ \\
QQP & $2_{0.06}$ & $2_{0.06}$& $2_{0.1}$ & $4_{0.09}$ & $3_{0.09}$ \\
RTE & $2_{0.07}$ & $2_{0.06}$& $2_{0.08}$ & $3_{0.1}$ & $2_{0.08}$ \\
MRPC & $2_{0.08}$ & $4_{0.1}$& $3_{0.09}$ & $7_{0.1}$ & $7_{0.1}$ \\
QNLI & $1_{0.08}$ & $2_{0.08}$& $3_{0.09}$ & $7_{0.09}$ & $3_{0.09}$ \\
\end{tabular}
\caption{\label{table:info_table} Approximate matrix rank of the correct class probability predictions of four model families on six GLUE task validation sets. We compute low-rank matrix approximations for 27 BERT-family predictions, 30 GPT-family predictions, 15 Instruction-tuned GPT-family predictions (IGPT), 15 OpenAI-family predictions (OAI), and all 87 model predictions (ALL) together. The approximations achieve very low mean absolute error (indicated by the subscripts), despite their ranks being considerably smaller than the number of models in each family.} 
\end{center}
\end{table}

\begin{table*}[t]
\centering
\footnotesize
\begin{tabular}{l@{\ }|cccccc|cc}
\toprule
& Random & Random &
Pretrained & Pretrained  & Fine-Tuned & Fine-Tuned & AP & AP \\
&& Mean   && Weighted &&  Weighted  & Weighted &  Predictor  \\
\midrule 
Datasets & Exact & Corr &  Exact & Corr & Exact & Corr & Corr & Exact \\
\midrule 
SST-2  & 0.685 & 0.705  & 0.734  & 0.725 & 0.730  & \textbf{0.787}  & \underline{0.757}  & 0.727 \\
QQP  & 0.669 & 0.678 & 0.189 & 0.233 & \underline{0.766} & \textbf{0.770} & 0.756 & 0.701 \\
RTE  & 0.366 & 0.308 & 0.143 & -0.052 & 0.354 & 0.275 & \textbf{0.483} & \underline{0.462} \\
QNLI   & 0.321 & \underline{0.331} & 0.192 & 0.294 & 0.127 & 0.144 & \textbf{0.439} & 0.303\\
MRPC  & 0.687 & 0.679 & 0.528 & 0.604 & 0.641 & 0.681 & \textbf{0.726} & \underline{0.716} \\
MNLI & 0.438 & 0.433 & 0.177 & 0.166 & 0.523 & 0.453 & \textbf{0.544} & \underline{0.517} \\
\midrule 
\textbf{Average} & 0.528 & 0.522 & 0.327 & 0.328 & 0.523 & 0.518 & \textbf{0.612} & \underline{0.571} \\

\bottomrule
\end{tabular}
\caption{Area Under the (Kendall's $\tau$) Correlation Curve from 1 to 30 points for ranking 77 language models at 6 General Language Understanding tasks (GLUE). We randomly select 10 models to be source models for the AP methods and rank the remaining 77 models, averaging over 100 randomized runs. AP Weighted and AP Predictor show significant gains, which are most dramatic at smaller evaluation sets (Figures \ref{fig:corrs-union-1} and \ref{fig:corrs-union-2}). "Exact" indicates the method generates a score that is intended to approximate the true aggregate performance directly, while "Corr" indicates the method generates a score intended only to rank correlate with true performance. The best score is \textbf{bolded} and second best score is \underline{underlined}. Table \ref{tab:AUCC_table_families} shows these results broken down by model family. } \label{tab:AUCC_table}

\end{table*}

\section{Sample-Efficient Model Evaluation}

\label{sec:results}

We now evaluate the anchor point techniques and
compare against baselines for 1) selecting representative evaluation subsets to rank models\footnote{We also evaluate anchor point techniques for ranking MMLU performance in Appendix \ref{apppendix:mmlu}} and 2) estimating model instance-level performance on held-out points in $\mathcal{D}_{eval}$ . 

\paragraph{Subset Selection Baselines and Metrics}
For baselines, we use uniform random sampling as well as K-Medoids sampling over the embedding spaces of a generic sentence encoder-- SentenceBERT \cite{sentencebert} -- and an encoder fine-tuned on the dataset-- the \emph{CLS} token of bert-base-uncased \cite{bert}. For these techniques, we use model performance on the selected points as the estimate of aggregate model performance on the entire dataset. We further consider variants of these baselines that consider model confidence. We propose mean model confidence in the correct class on randomly selected points as one baseline. We also propose to produce a score estimate for each model by computing a weighted average of the correct class probabilities assigned to each selected point. The weights are proportional to the size of each selected point's cluster and sum to 1, akin to Anchor Points Weighted.

To assess the ranking performance of each technique, we compute the Kendall rank correlation coefficient (Kendall's $\tau$) for a range of anchor point set sizes and report the resulting \textbf{Area Under the Correlation Curve} (AUCC) from 1 to a maximum budget $B$. We then normalize AUCC by dividing by the best possible area. In the experiments we use $B=30$, the small data regime for which it is reasonable for practitioners to manually inspect predictions. 

\paragraph{Prediction Estimation Baselines and Metrics}
We consider two naive baselines: 1) using the instance-level predictions of a randomly selected source model on $\mathcal{D}_{eval}$ as the estimate of each target model's prediction and 2) using the mean prediction of all source models on each point in $\mathcal{D}_{eval}$ as the estimate. We also propose Nearest Source Neighbor, a strategy where each target model's predictions on $k$ randomly selected points are compared to the predictions of all source models on the same points. We select the source model with the most similar predictions on these $k$ points, as measured with $L1$-distance. The predictions of this source model on all other points in $\mathcal{D}_{eval}$ are then used as the estimate of the target model's predictions.

To measure the performance of each technique, we use \emph{agreement}-- the percentage of points for which the estimation technique assigns the highest probability to the same class as the target model.

\paragraph{Anchor Points Show Competitive Performance at Ranking Models with Small Evaluation Sets}

Table \ref{tab:AUCC_table} shows the Kendal's $\tau$ AUCC of the methods for ranking 77 models belonging to all model families. These curves are shown in Figures \ref{fig:corrs-union-1} and \ref{fig:corrs-union-2}. Table \ref{tab:AUCC_table_families} shows performance within each model family. Overall, we observe that Anchor Points Weighted outperforms random selection in 29 of 30 settings and Anchor Points Predictor outperforms random selection in 27 of 30 settings. Anchor Points Weighted proves to be the strongest among all techniques, followed by Anchor Points Predictor. Very small anchor point sets achieve surprisingly strong correlation, serving as reliable evaluation sets that can easily be inspected by eye (e.g. Figure \ref{fig:sst2-example-AP-set}). Moreover, we still observe these gains when source models are freely-available while all target models are closed-source (Table \ref{tab:cheap-source-models}).

\paragraph{Anchor Point Weighted Correlations Are Reliable, Unlike Strong Contenders}

 We observe that Anchor Point Weighted (APW) performance tends to be more reliable across evaluation set sizes and test settings than APP as well as the baselines. We suspect APW is more reliable than APP because it makes a weaker assumption that source model predictions simply follow similar correlations as target model predictions rather than closely matching the exact regression line fitted to the source model predictions. This is corroborated by results in Appendix \ref{apppendix:mmlu} with the MMLU dataset, where APW still demonstrates superior performance over baselines despite a small number of source models, but APP performance worsens.

 We observe inconsistent and sometimes erratic behavior in the embedding-based baselines. Despite Fine-tuned and Fine-tuned Weighted baselines collectively achieving the best performance in 10 settings, these baselines prove to be inconsistent: Fine-tuned and Fine-tuned Weighted perform worse than random in 12 and 10 settings, respectively. We suspect this is because language model embedding spaces are not always smooth and are likely to be inconsistent across models.

\begin{table*}[t]
\centering
\footnotesize
\begin{tabular}{l@{\ }c|cccccc}
\toprule
&  & Random Source & Mean Source &
 \multicolumn{2}{c}{Nearest Source Neighbor} &
 \multicolumn{2}{c}{Anchor Points}  \\ 
\midrule 
Dataset & $|\mathcal{D}_{eval}| $ & $B$ = 0 & $B$ = 0 & $B$ = 10 & $B$ = $|\mathcal{D}_{eval}|$ & $B$ = 10 & $B$ = 100  \\
\midrule 
SST-2  & 872 & $0.63_{0.23}$  & $0.74_{0.17}$  & $0.83_{0.16}$ & $\mathbf{0.84_{0.15}}$  & $0.83_{0.12}$ & $\mathbf{0.84_{0.11}}$    \\
QQP  & 6000 & $0.54_{0.28}$ & $0.59_{0.22}$ & $0.84_{0.15}$ & $\mathbf{0.85_{0.14}}$ & $0.81_{0.13}$ & $0.83_{0.14}$  \\
RTE  & 277 & $0.53_{0.32}$ & $0.58_{0.31}$ & $0.80_{0.21}$ & $0.81_{0.20}$ & $0.83_{0.17}$ & $\mathbf{0.85_{0.16}}$   \\
QNLI  & 5463 & $0.60_{0.26}$ & $0.65_{0.23}$ & $0.78_{0.19}$ & $\mathbf{0.79_{0.19}}$ & $0.76_{0.17}$ & $0.75_{0.15}$ \\
MRPC  & 408 & $0.53_{0.26}$  & $0.56_{0.20}$ & $0.80_{0.15}$ & $0.82_{0.14}$ & $0.79_{0.14}$ & $\mathbf{0.84_{0.11}}$    \\
MNLI  & 6000 & $0.40_{0.30}$ & $0.46_{0.28}$ &$0.71_{0.26}$ & $0.72_{0.25}$ & $\mathbf{0.79_{0.20}}$ & $0.76_{0.20}$  \\
\midrule 
\textbf{Average} & -- & 0.54 & 0.60 & 0.79 & \textbf{0.81} & 0.80 & \textbf{0.81}  \\
\bottomrule
\end{tabular}
\caption{Agreement of true target model predictions and estimated target model predictions for anchor points and baselines, where $B$ indicates the number of points the target models are evaluated on. We randomly select 10 models to be source models and estimate the predictions of the remaining 77 models, averaging over 100 randomized runs. Anchor Points and Nearest Source Neighbor surpass the naive baselines by a large margin. Standard deviations (subscripts) remain large, indicating the presence of outlier models. The $B = |\mathcal{D}_{eval}|$ column is shown only for reference to highlight that Anchor Points can surpass the upper bound of Nearest Source Neighbor performance. Note that standard errors can be computed by dividing the standard deviations by 10. The best score is \textbf{bolded}.} \label{tab:agreement_table}
\end{table*}

\paragraph{Anchor Points Efficiently Estimate Model Predictions}
The Anchor Point Predictor can estimate model performance at the instance-level over the entire dataset. Table \ref{tab:agreement_table} shows the average agreement between Anchor Point estimated instance-level model predictions and true model predictions using various numbers of anchor points and randomly selected source models. The estimates of a small number of anchor points achieve high average agreement with true model predictions. However, the source model correlations do not generalize well to all target models, resulting in lower agreement on these outlier models and thus large standard deviations in Table \ref{tab:agreement_table}. As a general rule of thumb, selecting a source model set that is diverse (e.g. comes from multiple families, see Figure \ref{fig:transfer-table}) results in the best generalization. Strategies for intelligently selecting source models are a promising direction to resolve poor anchor point generalization, which we discuss further in Appendix \ref{appendix:generalization}. 

\paragraph{Nearest Source Neighbor is a Strong Baseline}

Nearest Source Neighbor achieves competitive performance at estimating model instance-level predictions. The agreement achieved by this technique is upper bounded by the agreement of each target model and its true nearest source model neighbor (indicated by the $B$ = |$\mathcal{D}_{eval}|$ column in Table \ref{tab:agreement_table}). Surprisingly, comparing just $B = 10$ random points closely approaches this bound, suggesting that the similarity of the predictions of different models can be approximated cheaply. Note that using $B = 100$ anchor points surpasses the upper bound of this baseline on 3 of the 6 datasets.

\section{Sample-Efficient Model Analysis}

We now highlight how anchor point maps provide fine-grained analysis of model generalization both within and across datasets. This is achieved in a sample-efficient manner using the Anchor Point Predictor. 

\paragraph{Anchor Point Maps Visualize Where Models Generalize}

Anchor Point Maps reveal the extent to which models learn distinct regions of dataset distributions. Model performance on each sample correlates with performance in the sample's neighborhood, allowing the Anchor Point Predictor to estimate where models will fail without evaluating the entire dataset. Figure \ref{fig:main-corrmap} shows the SST-2 validation set mapped by the predictions of 60 source models and then colored by estimated Falcon-7B predictions using thirty anchor points. These estimated predictions are quite faithful to the model's true predictions (achieving 0.09 MAE and 92\% agreement), revealing regions where the model is weak. This localization of model behavior starkly contrasts with typical language embedding spaces, where model performance tends to be non-localized (Figure \ref{fig:embedding-comparison}). However, like embedding spaces, related examples naturally cluster together.

\paragraph{Anchor Points Predict Diverse Model Behavior}

Figure \ref{fig:qqp-corrmap-comparison} in the Appendix shows four Anchor Point Maps of 1000 QQP points, comparing the true and estimated predictions of deberta-v3-base and text-davinci-003 using 30 anchor points. The two models are weak in distinct regions of the dataset. Despite deberta-v3-base achieving stronger performance overall, it fails in a region of the negative class cluster where text-davinci-003 is mostly correct. This is perhaps related to "no-free lunch" theorems in statistical learning theory: different models have biases that will lend them to perform the best on different inputs \cite{no-free-lunch}, which is obscured by single-number evaluation metrics. The fact that the same set of anchor points can generalize to models with diverse behaviors highlights that the same underlying predictive correlations can engender these different behaviors. We explore anchor point generalization further in Appendix \ref{appendix:generalization}.

\begin{figure*}[t]
  \centering
  \subfigure[Anchor Point Map of 6 Combined MMLU datasets. The map is computed using the predictions of 13 source models. We color each point according to the dataset to which it belongs. We observe that all datasets overlap, but points tend to cluster with other points from the same dataset or a related dataset. \label{fig:mmlu_map}] {\includegraphics[scale=0.35]{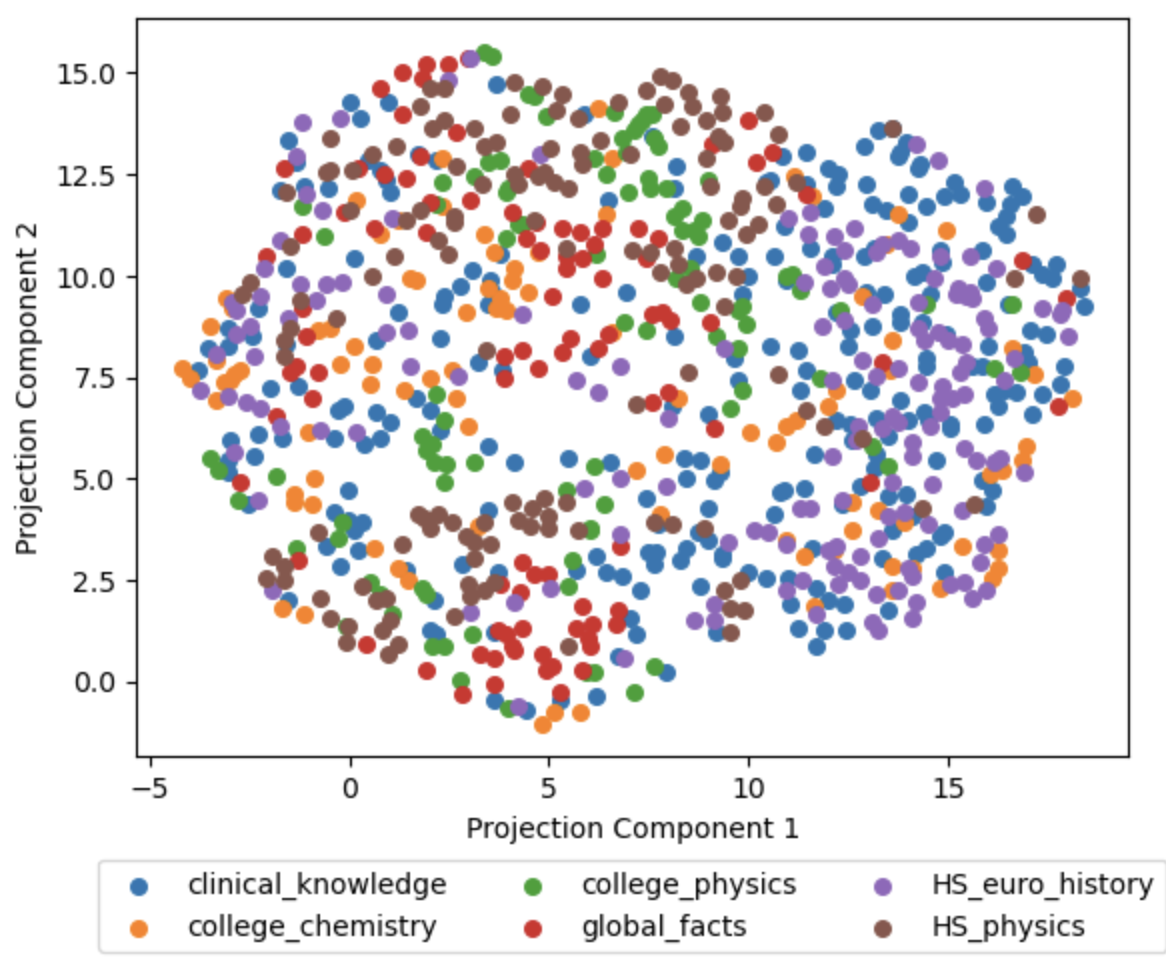}} \quad
  \subfigure[For each question type (i.e dataset), we show the distribution of question types of its 10 nearest neighbors (on average). For example, the top row shows that 49\% of the nearest neighbors of clinical knowledge (CK) questions tend to also be clinical knowledge, while 19\% tend to be high school European history (HSEH). Labels: CK = clinical knowledge, CC = college chemistry, CP = college physics, GF = global facts, HSES = high school European history, HSP = high school physics. \label{fig:mmlu_NN}]{\includegraphics[scale=0.395]{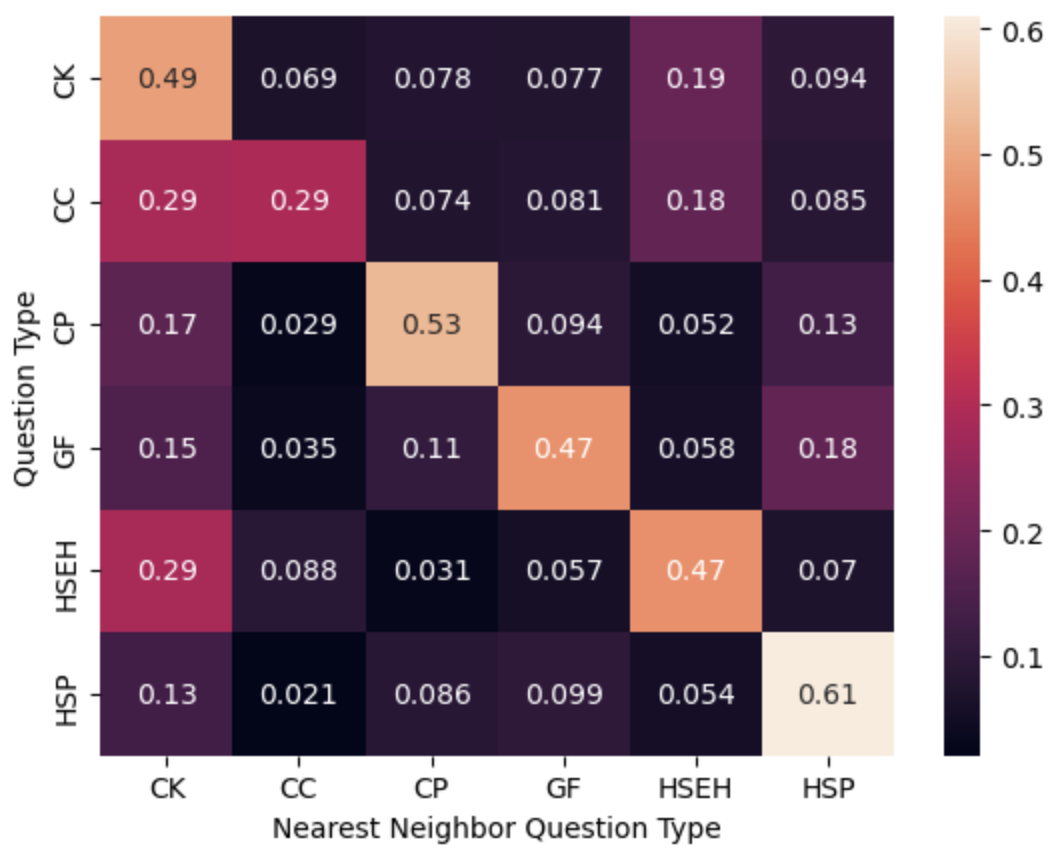}}

  \caption{Analysing Patterns in Model Knowledge using MMLU Anchor Point Map.}
  
\end{figure*}

\paragraph{Anchor Point Maps Reveal Patterns in Model Knowledge}

To explore the characteristics of anchor points beyond basic language understanding tasks, we generate an anchor point map (Figure \ref{fig:mmlu_map}) using the predictions of 14 LLMs on 6 MMLU datasets \cite{mmlu}. Each dataset contains multiple choice questions from a distinct domain, requiring substantial real-world knowledge to answer. We observe overlap of all domains in the anchor point map, but there is a remarkable amount of structure: models tend to have correlated performance on groups of questions from the same domain as well as related domains. This is quantified is Figure \ref{fig:mmlu_NN}, which shows the distribution of domains of each question type's 10 nearest neighbors. On average, around 60\% of the nearest neighbors of a high school physics question are also from the high school physics domain. In contrast, a nearest neighbor of a college chemistry question is equally likely to be about college chemistry or clinical knowledge. This suggests that models with strong knowledge of college chemistry are likely to have strong knowledge of clinical knowledge as well. This is perhaps due to frequent co-occurrence of these subjects in pre-training data.

\paragraph{Anchor Points Capture Model Performance, Not Language Semantics}
Notably, Figure \ref{fig:mmlu_NN} reveals that high school physics and college physics questions are often not nearest neighbors, despite these datasets having the largest vocabulary overlap (21\%) of any pair within the 6 MMLU datasets (Figure \ref{fig:vocab-overlap}). This highlights that the semantic similarity of two questions is often a poor proxy for similar model performance on the questions, explaining why embedding-based approaches to sample-efficient model evaluation tend to fail. Anchor Point Selection is more akin to test-distribution aware active learning, where points are selected based on the information they provide about future model predictions \cite{mackay, test-aware}.

\section{Limitations}
\label{sec:limitations}

\paragraph{Anchor Points Show Diminishing Returns}

We discuss in Appendix \ref{appendix:generalization} that APW correlation begins to plateau as the evaluation set size grows to 100 points due to an inherent upper bound. We also discuss that for large evaluation sets, anchor points can underperform random selection. More work is needed to develop methods that are robust to evaluation set size.

\paragraph{Anchor Points Do Not Always Generalize}
In order for anchor points and anchor point maps to generalize, the predictive trends of source models must be consistent with that of target models. While Table \ref{tab:AUCC_table_families} shows that anchor points often transfer both within and across model families, the correlations of randomly selected source models do not necessarily generalize to all target models (e.g. Figure \ref{fig:curie-failure}). We currently lack rigorous theory to guide source model selection for evaluation of a given set of target models. 

\paragraph{Anchor Points Can be Further Optimized}
In its current form, Anchor Point Maps require computing correlations between all pairs of points in the dataset. This is computationally expensive for huge datasets. Multi-dimensional scaling with missing values or correlation matrix completion could dramatically decrease these costs.

\section{Conclusion and Future Work}

We present Anchor Point Selection, a technique that finds maximally-representative subsets from large datasets which can be used to efficiently rank models and estimate model behavior over the entire dataset. We also present Anchor Point Maps, a tool to visualize how well models generalize across various subsets of datasets.

Important future work includes developing better theory underlying when models share predictive correlations and designing intelligent source model selection strategies to ensure anchor point generalization to desired sets of target models.

The fact that diverse language models make highly correlated predictions on many pairs of examples suggests that modern benchmarks contain many redundant examples. An interesting future direction would be to use predictive correlations to guide benchmark development by selecting fewer redundant examples, which would ideally lead to more diverse and difficult benchmarks.

The potential to extend Anchor Point Maps and Anchor Point Selection to tasks beyond language classification is exciting. Perhaps other continuous performance metrics are strongly correlated between examples, such as BLEU scores or the rewards given by a reward model. Exploiting such phenomena could allow a dramatic reduction in the number of examples that modern language models must be evaluated on, leading to reduced costs for model development.

\section*{Ethics Statement}
We aim to encourage the research community to consider how modern NLP models can be evaluated more efficiently. While this endeavor could ideally reduce the compute required to develop models, minimizing evaluation set size poses the risk of excluding minority subsets of the data distribution. This could harm model generalization and lead to reduced performance in rarer use cases, e.g. on low-resource languages. As an empirical example, observing where Figures \ref{fig:falcon-est} and \ref{fig:falcon-true} differ reveals a small subset of points in sparser regions that are poorly captured by the anchor points. We find that examples in this region remain poorly captured across random seeds and anchor point set sizes, highlighting a systematic lack of generalization to this minority subset.

To mitigate the risks of our work, we 1) emphasize that our ultimate goal is not to replace large benchmarks but rather provide cheaper signal about model benchmark performance and 2) remain candid about the limitations of our approach to lay the foundation for future work in efficient, robust model evaluation strategies.

\nocite{*}
\bibliography{anthology,custom}
\bibliographystyle{acl_natbib}

\newpage
\appendix

\section{Anchor Points Generalization Analysis}

\begin{table}[t]
\centering
\footnotesize
\begin{tabular}{l@{\ }|ccc}
\toprule
& Random & AP Weighted & AP Predictor  \\
\midrule 
SST-2  & 0.668 & \textbf{0.744}  & 0.714  \\
QQP  & 0.606 & \textbf{0.729}  & 0.701 \\
RTE & 0.335 & 0.372  & \textbf{0.453} \\
QNLI & 0.449 & \textbf{0.514}  & 0.312 \\
MRPC  & 0.648 & \textbf{0.711}  & 0.668  \\
MNLI & 0.302 & 0.366  & \textbf{0.411}  \\
\midrule 
\textbf{Average} &0.502 & \textbf{0.573} & 0.543 \\

\bottomrule
\end{tabular}
\caption{Area Under the (Kendall's $\tau$) Correlation Curve from 1 to 30 points for ranking the accuracy of 15 OPENAI model-prompt pairs on 6 GLUE datasets. We randomly select 10 free, open-source models to be source models for the AP methods and rank the remaining 7 models, averaging over 100 randomized runs. The best score is \textbf{bolded}. We observe that APW and APP have an average improvement over random by 0.07 and 0.04 AUCC respectively. This highlights that the anchor point methods can reduce the cost of ranking expensive target models with minimal source model costs.} \label{tab:cheap-source-models}
\end{table}

\begin{figure}[]
\includegraphics[scale=0.30]{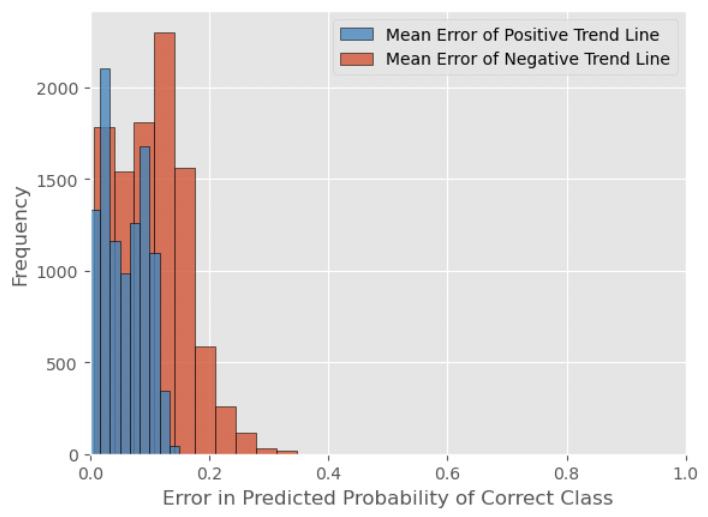}

\caption{\label{fig:outlier_analysis} Mean Absolute Error Distribution for regression lines fit to source model predictions with moderate-to-strong fit ($R^2$ > 0.64) and used to predict target model predictions. Generalization is worse for trends with negative slopes.}

\end{figure}

We perform further investigations to assess under what conditions the predictive correlations of source models generalize well to target models. Figure \ref{fig:outlier_analysis} shows how well strong positive and strong negative trends in source model predictions generalize to target model predictions. We observe that positive trends generalize more reliably than negative trends. Negative trends suggest that models struggle to correctly label both of two examples correctly: greater confidence in the correct class in one example correlates with lesser confidence in the correct class of the other example. We hypothesize that this characteristic is a weakness of some models that does not generalize to stronger models, leading to poor generalization. Note that Equation \ref{eq:objective-2} selects points with strong positive trends, avoiding this issue.

\begin{figure*}[t]
  \centering
  \subfigure[Pairwise Correlation Matrix of 87 Model-Prompt Pairs Correct Class Confidences on 50 SST-2 Examples. Rich structure indicates the relatedness of various examples. Notably, a dark bar at index 13 indicates an example that tends to negatively correlate with all other examples.\label{fig:sst2-corrmat}] {\includegraphics[scale=0.32]{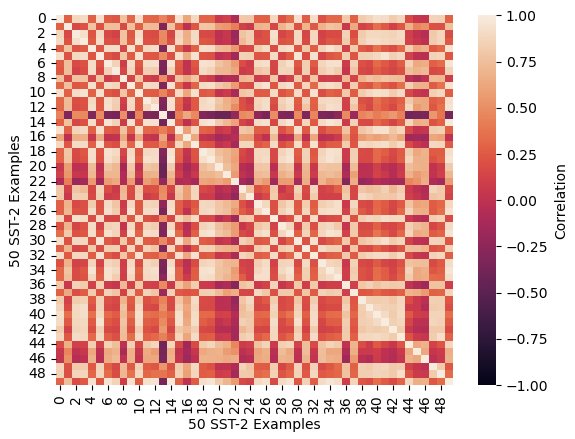}} \quad
  \subfigure[Correct Class Confidences of 87 Models on Two Selected SST-2 sentence pair examples. Example 1 (x-axis) corresponds to the dark bar in \ref{fig:sst2-corrmat}, a highly contentious example. Prompted models follow a negative trend while BERT models follow no trend, highlighting the distinct behavior of fine-tuned vs. prompted models.  \label{fig:contentious}]{\includegraphics[scale=0.32]{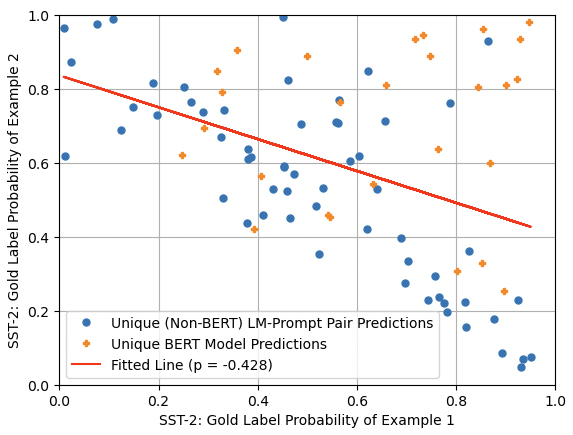}}
    
  \caption{Predictive Correlations at the Instance-Level Across Language Models: SST-2 Case Study}
  \label{fig:sst2-case-study}
  
\end{figure*}

Figure \ref{fig:sst2-case-study} shows the correlation matrix of 50 randomly selected SST-2 points, computed across the predictions of all 87 models. We observe that nearly all correlations are positive, with a notable exception. This exception is a highly contentious example: "we root for ( clara and paul ) , even like them , though perhaps it 's an emotion closer to pity, " labeled as positive sentiment. We observe in Figure \ref{fig:contentious} that prompted language models tend to follow a negative trend between this example and others, while fine-tuned BERT models do not. This suggests a distinction in how fine-tuned and prompted models fit the task distribution. Notably, it highlights that different model families may have predictive correlations that do not generalize to other families. Further analysis of these properties could lead to the design of intelligent source selection schemes for evaluating a given set of target models.

\begin{figure*}[t]
  \centering
  \subfigure[QNLI] {\includegraphics[scale=0.6]{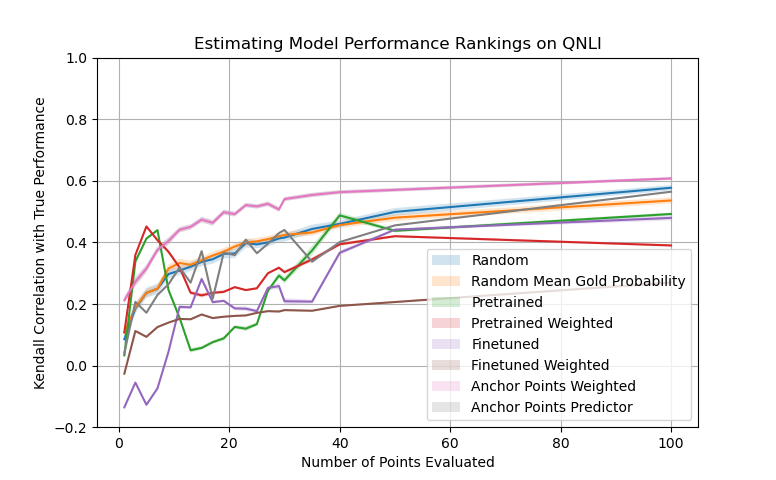}} \quad
  \subfigure[MRPC]{\includegraphics[scale=0.6]{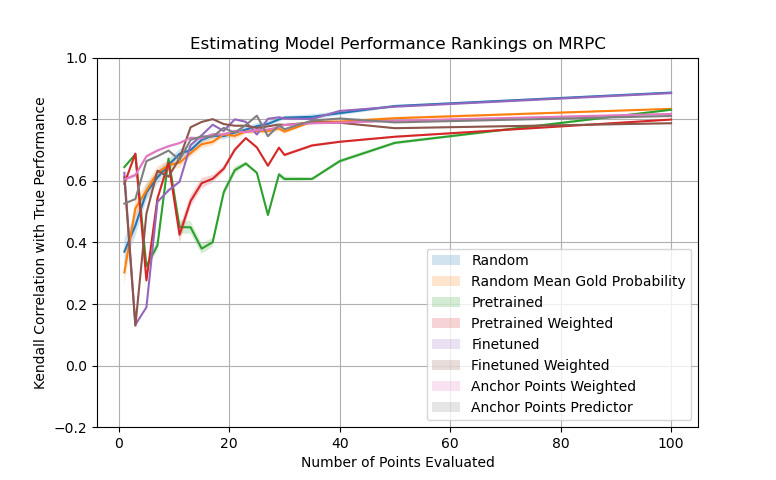}}
  \subfigure[MNLI]{\includegraphics[scale=0.6]{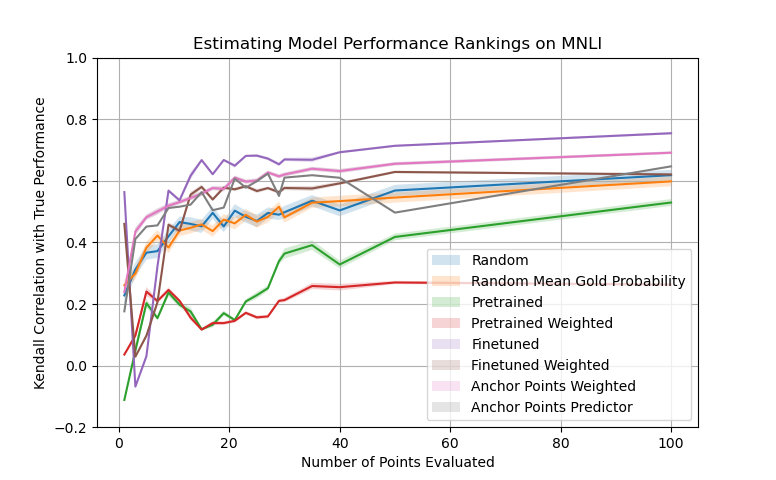}} \quad
  \caption{Kendall Tau Rank Correlation between Model Rankings on Small Evaluation Sets and Full Validation Sets for QNLI, MRPC, and MNLI. We rank 77 language models belonging to all model families using various evaluation selection techniques. Anchor Points are fit to 10 source model predictions. Each point is the mean of 100 runs with randomized source and target models. Shading indicated standard error. Anchor Points Weighted achieves the most reliable performance overall at low evaluation set sizes.}
  \label{fig:corrs-union-1}
  
\end{figure*}

\begin{figure*}[t]
  \centering
  \subfigure[QQP] {\includegraphics[scale=0.6]{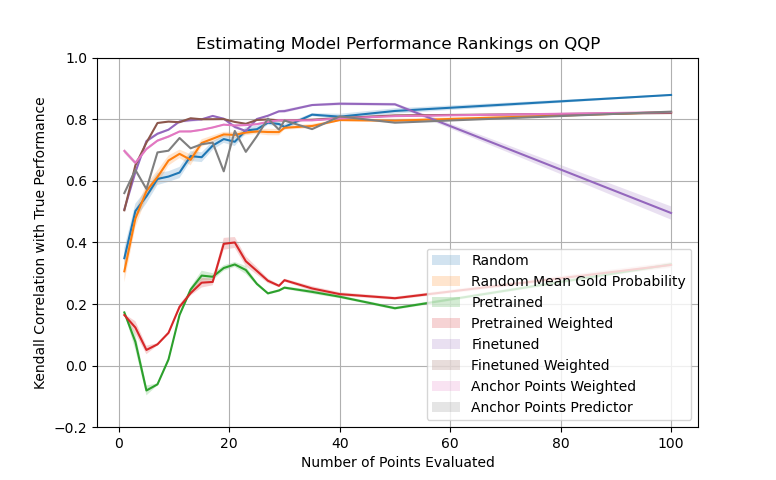}} \quad
  \subfigure[RTE]{\includegraphics[scale=0.6]{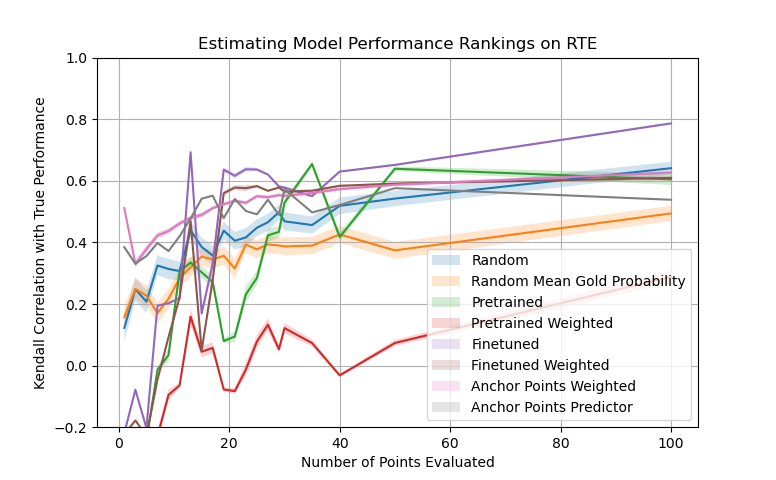}}
  \subfigure[SST-2]{\includegraphics[scale=0.6]{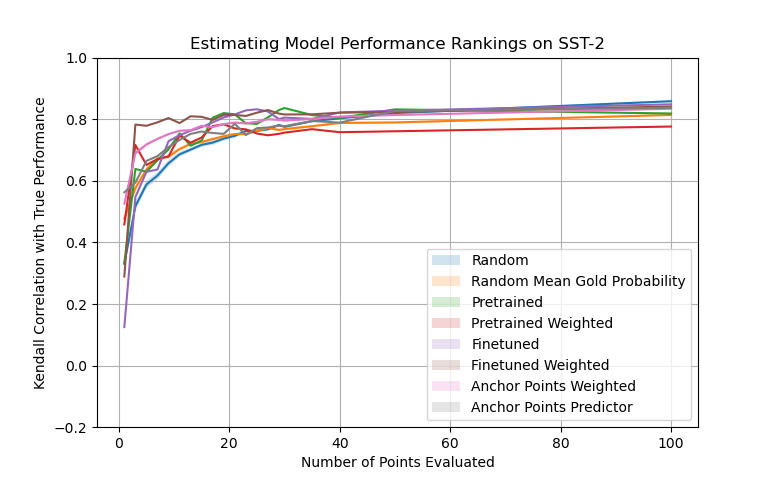}} \quad
  \caption{Kendall Tau Rank Correlation between Model Rankings on Small Evaluation Sets and Full Validation Sets for QQP, RTE, and SST-2. We rank 77 language models belonging to all model families using various evaluation selection techniques. Anchor Points are fit to 10 source model predictions. Each point is the mean of 100 runs with randomized source and target models. Shading indicated standard error. Anchor Points Weighted achieves the most reliable performance overall at low evaluation set sizes.}
  \label{fig:corrs-union-2}
  
\end{figure*}

Figures \ref{fig:corrs-union-1} and \ref{fig:corrs-union-2} show the Kendall tau correlation curves for the various model ranking methods. For many settings, APW correlation begins to plateau as the evaluation set size grows to 100 points. For 19 of the 30 settings, the random selection curve eventually reaches the APW curve at some evaluation set size smaller than 100 points. This is natural: a randomly-selected evaluation set is sufficiently representative when sufficiently large. These diminishing returns occur more quickly for simpler datasets (e.g. SST-2) and less so for more complex datasets (e.g. MNLI), reflecting that more complex distributions require larger evaluation sets to be well-represented. However, it is surprising that random selection eventually \emph{surpasses} the APW curve in 5 of these 19 settings. This is a result of the APW curve having a tighter upper bound than random selection: in the limit where $|\mathcal{X}_{acq}| = |\mathcal{X}_{eval}|$, instance-level confidence in the correct class averaged over the entire dataset does not perfectly correlate with accuracy over the entire dataset. This upper bound does not exist for APP.
\label{appendix:generalization}

\begin{figure*}[]
  \centering
  
  \subfigure[QQP Anchor Point Map Colored with Estimated deberta-v3-base predictions. The estimates achieve an MAE of 0.11 and agreement of 89\%. \label{fig:deberta-est}]{\includegraphics[scale=0.35]{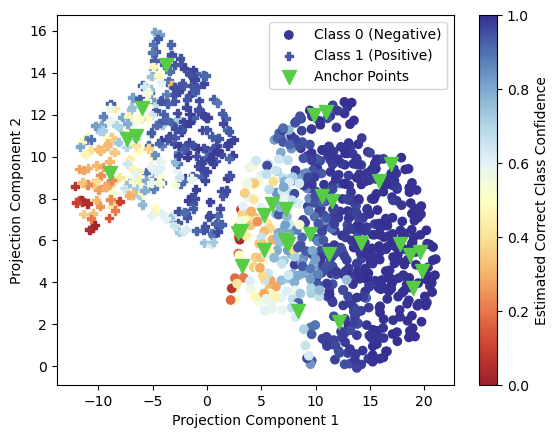}} \quad
  \subfigure[QQP Anchor Point Map Colored with Estimated text-davinci-003 (Prompt 1) predictions. The estimates achieve an MAE of 0.16 and agreement of 76\%.\label{fig:davinci-est}]{\includegraphics[scale=0.35]{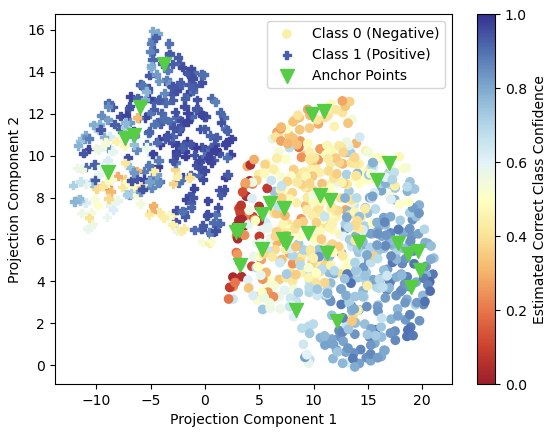}}
  \subfigure[QQP Anchor Point Map Colored with True deberta-v3-base predictions.\label{fig:deberta-true}] {\includegraphics[scale=0.35]{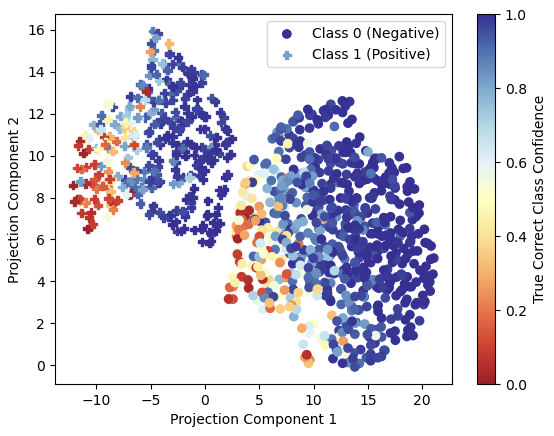}} \quad
  \subfigure[QQP Anchor Point Map Colored with True text-davinci-003 (Prompt 1) predictions. \label{fig:davinci-true}]{\includegraphics[scale=0.35]{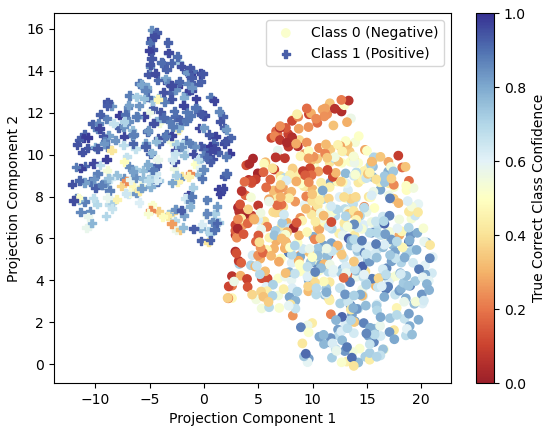}} 
    
  \caption{Anchor Point Map for 1000 QQP examples. The map is computed using the predictions of 60 randomly-selected source models. We then estimate the predictions of the two held-out target models, deberta-v3-base and text-davinci-003, by evaluating each on 30 anchor points. We color the remaining 970 test points in \ref{fig:deberta-est} and \ref{fig:davinci-est} with these estimates. Finally, we color maps \ref{fig:deberta-true} and \ref{fig:davinci-true} with the true target model predictions. We observe that the estimated predictions achieve low MAE and high agreement with the true predictions.} 
  \label{fig:qqp-corrmap-comparison}
  
\end{figure*}

\begin{figure}[]
  \centering
    {\includegraphics[scale=0.37]{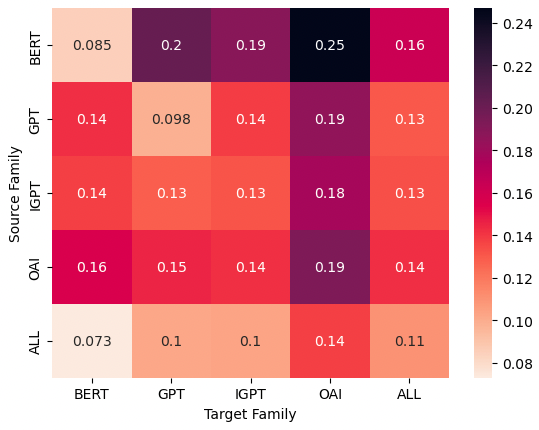}} 
  \caption{Transfer table showing the mean absolute error of trend lines fit to the instance-level predictions of one model family and used to estimate the instance-level predictions of another family. Along the diagonal, we partition the family into source and target halves randomly. Results are averaged over 1000 points pairs from each of the 6 GLUE tasks. Point pairs are randomly selected from all pairs having a Pearson correlation greater than +0.8 within the source model predictions.}
  \label{fig:transfer-table}
  
\end{figure}

\begin{figure*}[]
  \centering
  
  \subfigure[SST-2 Anchor Point Map Colored with Estimated Falcon-7b (Prompt 3) Predictions. The estimates achieve an MAE of 0.09 and agreement of 92\%. \label{fig:falcon-est}]{\includegraphics[scale=0.32]{emnlp2023-latex/figures/final_corrmaps/final_corrmap_main_big.png}} \quad
  \subfigure[SST-2 Anchor Point Map Colored with True Falcon-7b (Prompt 3) Predictions.\label{fig:falcon-true}]{\includegraphics[scale=0.32]{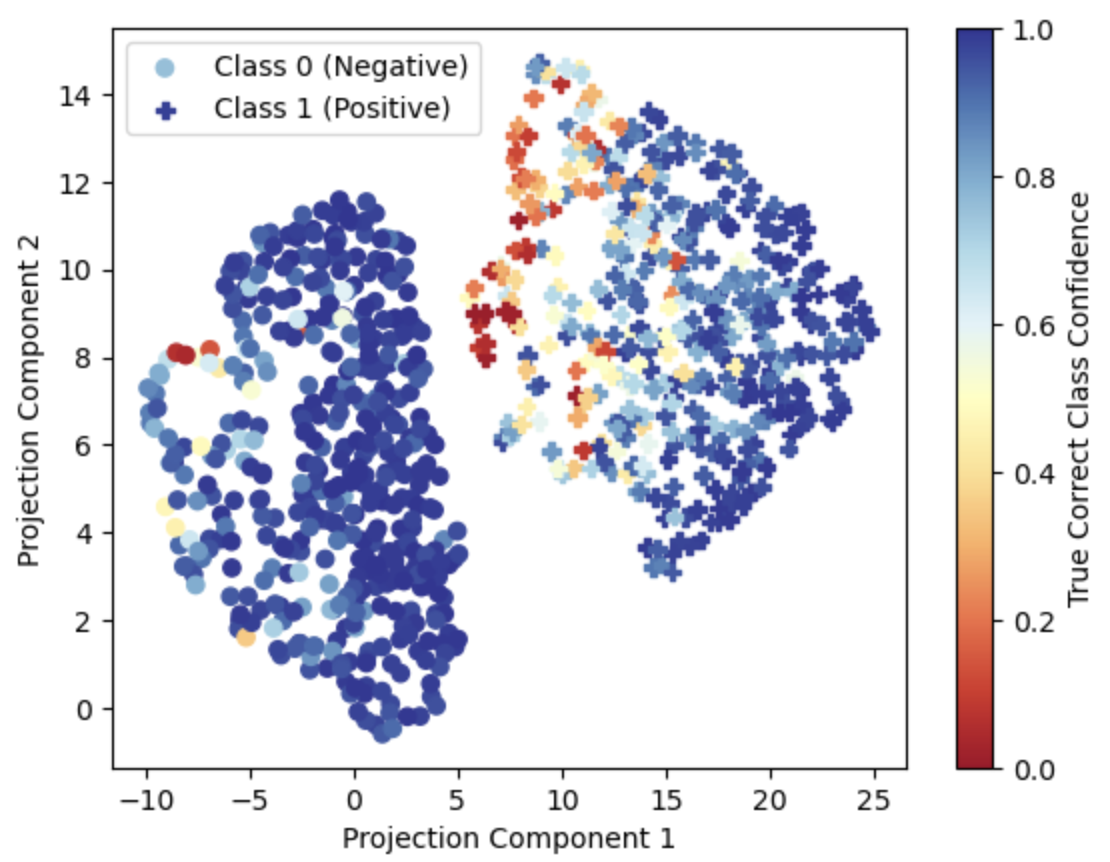}}

  \caption{SST-2 Validation Set Anchor Point Map. The locations of all 872 points are learned using the predictions of 60 randomly-selected source models on SST-2. We then evaluate Falcon-7b (held-out) on 30 anchor points, shown by green triangles in \ref{fig:falcon-est}. Next, the model's predictions on only these anchor points is used to estimate the models' predictions on the remaining 842 points,  with a mean absolute error of 0.09 and 92\% agreement with the true predictions. The estimates are also shown in \ref{fig:falcon-est}. achieving 0.09 MAE and 92\% agreement with the models' true predictions. Finally, we color the Anchor Point Map with Falcon-7B's true predictions in \ref{fig:falcon-true}. The estimated predictions successfully identify regions where the model is weak (red regions).} \ 
  \label{fig:sst2-corrmap-comparison}
  
\end{figure*}

\begin{figure*}[]
  \centering
  
  \subfigure[Slope-Intercept Plot of trend lines fit across the predictions of 10 source models randomly selected from all families on 200 randomly-selected MNLI points. \label{fig:mnli_10_slopeint}]{\includegraphics[scale=0.35]{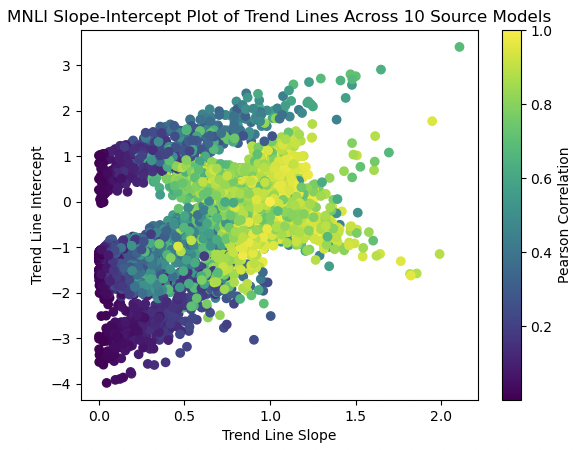}} \quad
  \subfigure[Slope-Intercept Plot of trend lines fit across the predictions of 10 source models randomly selected from all families on 200 randomly-selected SST-2 points.\label{fig:sst2_10_slopeint}] {\includegraphics[scale=0.35]{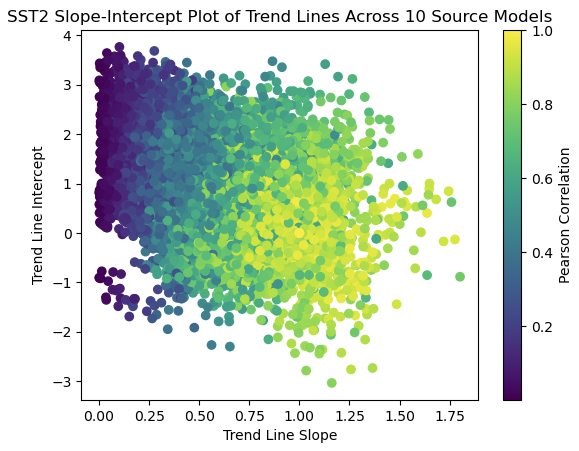}} \quad
    \subfigure[Slope-Intercept Plot of trend lines fit across the predictions of 87 source models from all families on 200 randomly-selected MNLI points.\label{fig:mnli_87_slopeint}]{\includegraphics[scale=0.35]{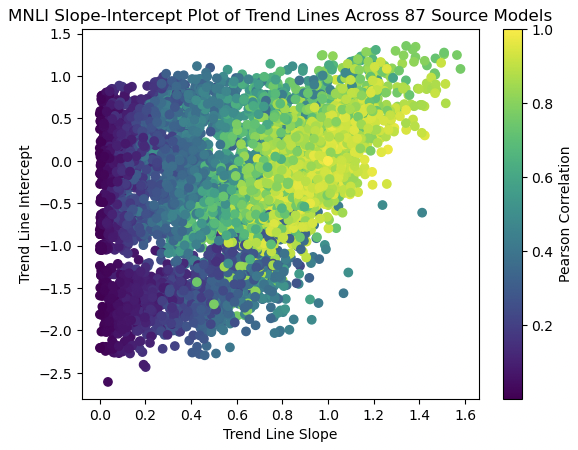}}
  \subfigure[Slope-Intercept Plot of trend lines fit across the predictions of 87 source models from all families on 200 randomly-selected SST-2 points. \label{fig:sst2_87_slopeint}]{\includegraphics[scale=0.35]{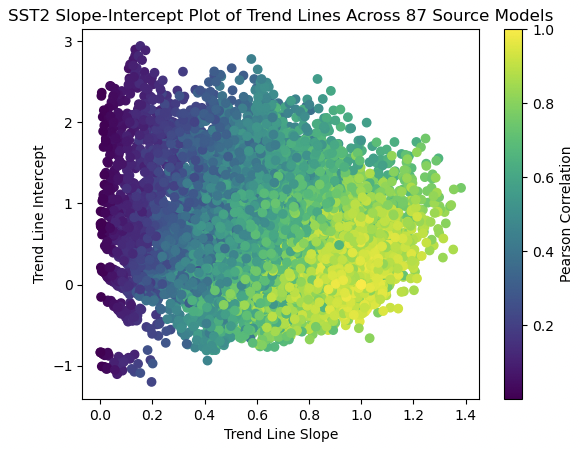}} 
    
  \caption{Slope-Intercept Plots of Trend Lines Fit Across the Predictions of Source Models on all pairs of 200 points randomly selected from MNLI and SST-2. Note that high Pearson correlations naturally emerge when trend lines have a slope near one and intercept near zero. This holds even when the number of source models is small (e.g. 10), suggesting that spuriously high correlations are not common.} 
  \label{fig:slope-intercept}
  
\end{figure*}

\begin{figure*}[]
  \centering
    {\includegraphics[scale=0.37]{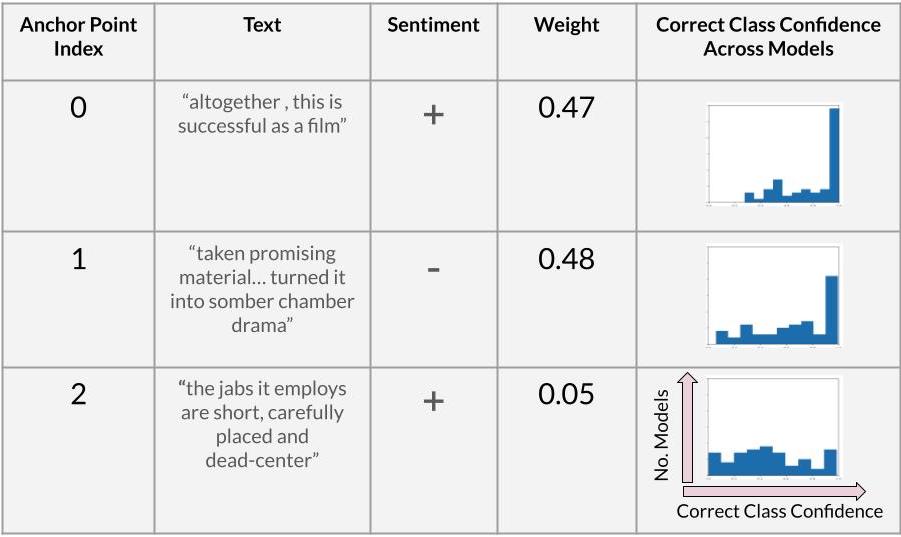} }
  \caption{Anchor point sets can be highly representative despite being small. This figure shows three SST-2 Anchor Points. The Anchor Point Weighted Score computed using these three points achieves a Kendall Tau correlation of 0.68 with true model performance on the SST-2 validation set (872 points), sufficient to identify the stronger performer of a randomly selected pair of models with 84\% probability. (Evaluating the pair on three randomly selected points would identify the stronger performer with 67\% probability). By observing the distribution of the correct class confidence of many models on these three points, we see that these points correspond to easy (0), moderate (1), and difficult (2) examples. }
  \label{fig:sst2-example-AP-set}
  
\end{figure*}

\begin{table*}[t]
\centering
\footnotesize
\begin{tabular}{l@{\ }|cccc|cccc}
\toprule
 &
 \multicolumn{4}{c|}{Nearest Source Neighbor} &
 \multicolumn{4}{c}{Anchor Points}  \\ 
\midrule 
Dataset  &  $N = 5$ &$N = 10$ & $N = 30$ & $N = 50$ & $N = 5$ & $N = 10$ & $N = 30$ & $N = 50$ \\
\midrule 
SST-2    & $0.78_{0.19}$  & $0.83_{0.16}$ & $0.86_{0.12}$  & $0.87_{0.12}$  & $0.78_{0.15}$ & $0.83_{0.12}$& $0.85_{0.11}$& $0.85_{0.12}$  \\
QQP  & $0.77_{0.21}$ & $0.84_{0.15}$ & $0.87_{0.13}$ & $0.88_{0.13}$ & $0.79_{0.15}$& $0.81_{0.13}$& $0.84_{0.11}$& $0.85_{0.12}$ \\
RTE   & $0.74_{0.24}$ & $0.80_{0.21}$ & $0.84_{0.18}$ & $0.86_{0.16}$ & $0.78_{0.19}$ & $0.83_{0.17}$& $0.80_{0.18}$& $0.87_{0.16}$ \\
QNLI   & $0.73_{0.22}$ & $0.78_{0.19}$ & $0.83_{0.17}$ & $0.83_{0.16}$ & $0.69_{0.17}$& $0.76_{0.17}$& $0.77_{0.17}$& $0.77_{0.16}$ \\
MRPC    & $0.77_{0.16}$ & $0.80_{0.15}$ & $0.82_{0.15}$ & $0.83_{0.15}$ &  $0.75_{0.15}$  & $0.79_{0.14}$& $0.85_{0.10}$& $0.81_{0.15}$ \\
MNLI   & $0.65_{0.27}$ &$0.71_{0.26}$ & $0.76_{0.24}$ & $0.78_{.24}$ & $0.76_{0.20}$  & $0.79_{0.20}$& $0.80_{0.20}$& $0.79_{0.20}$\\
\midrule 
\textbf{Average}  & 0.74 & 0.79 & 0.83 & 0.84 & 0.76 & 0.80& 0.82& 0.82  \\

\bottomrule
\end{tabular}
\caption{Anchor Point Agreement for Various Source Model Set Sizes on GLUE. We assess Nearest Source Neighbor and Anchor Points using $B=10$ points in all settings. We observe that Anchor Points are the stronger performer for smaller source model sets, but are surpassed by Nearest Source Neighbor for larger sets. Note that anchor points are stronger across the board for MNLI and weaker across the board for QNLI, indicating that the nature of the dataset plays a large role. }
\label{tab:ab2}
\end{table*}

\begin{table*}[t]
\centering
\footnotesize
\begin{tabular}{l@{\ }|cccccc|cc}
\toprule
& Random & Random &
Pretrained & Pretrained  & BGE & BGE Weighted & AP & AP \\
&& Mean   && Weighted &&  Weighted  & Weighted &  Predictor  \\
\midrule 
Datasets & Exact & Corr &  Exact & Corr & Exact & Corr & Corr & Exact \\
\midrule 
Clinical Knowledge  & 0.532 & 0.638  & 0.593  & \underline{0.705} & 0.542  & 0.491  & \textbf{0.711}  & 0.640 \\
College Chemistry  & 0.390 & 0.445  & 0.148 & 0.332 & 0.384  & \underline{0.472}  & \textbf{0.477}  & 0.267 \\
College Physics & 0.251 & 0.280  &\underline{0.310}  & 0.206 & 0.077 & 0.30  & \textbf{0.343}  & 0.130 \\
Global Facts & \underline{0.251} & 0.116  & 0.117  & 0.118 & \textbf{0.384}  & 0.109  & 0.151  & 0.052 \\
HS Euro. History  & 0.596 & 0.656  & 0.612  & \underline{0.678} & 0.677  & \textbf{0.741}  & 0.667  & 0.620 \\
HS Physics & 0.245 & 0.215  & 0.307  & \textbf{0.339} & -0.014  & -0.27  & \underline{0.324}  & 0.307 \\
\midrule 
\textbf{Average} &0.378 & 0.392 & 0.350 & \underline{0.406} & 0.341 & 0.262 & \textbf{0.445} & 0.336 \\

\bottomrule
\end{tabular}
\caption{Area Under the (Kendall's $\tau$) Correlation Curve from 1 to 30 points for ranking the accuracy of 7 language models on 6 MMLU datasets. We randomly select 7 models (see Table \ref{tab:mmlu-models}) to be source models for the AP methods and rank the remaining 7 models, averaging over 100 randomized runs. "Exact" indicates the method generates a score that is intended to approximate the true aggregate performance directly, while "Corr" indicates the method generates a score intended only to rank correlate with true performance. The best score is \textbf{bolded} and second best score is \underline{underlined}. } \label{tab:AUCC_table_MMLU}
\end{table*}

\begin{figure*}[]
  \centering
  
  \subfigure[1000 QQP Points Embedded with SentenceBERT and colored with the correct class confidence of bert-large-cased. \label{fig:pretrained_emb}]{\includegraphics[scale=0.32]{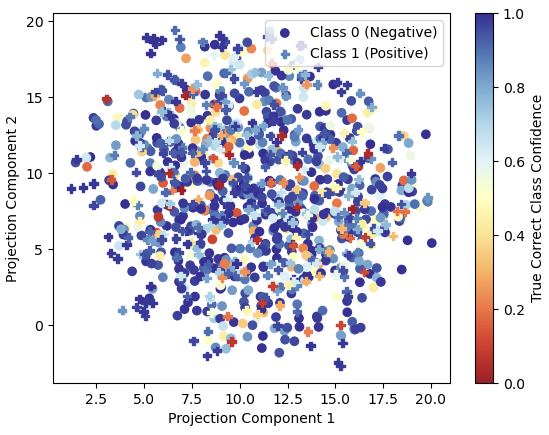}} \quad
  \subfigure[1000 QQP Points Embedded with the CLS token of bert-base-uncased (fine-tuned on QQP). Points are colored with the correct class confidence of bert-large-cased.\label{fig:finetuned_emb}]{\includegraphics[scale=0.32]{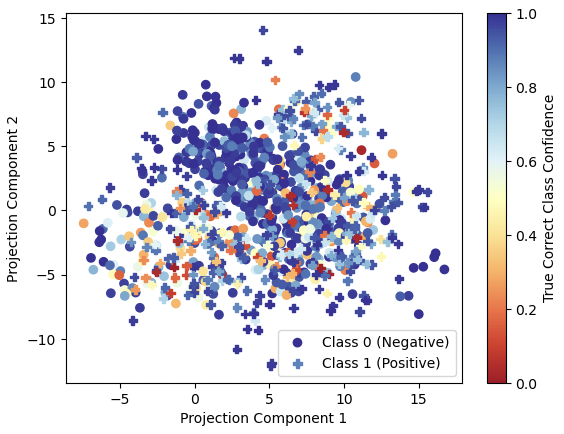}}
  \quad
  \subfigure[1000 QQP Points Visualized with the Anchor Point Map using 60 source models and colored with the correct class confidence of bert-large-cased. \label{fig:corrmap_emb}]{\includegraphics[scale=0.32]{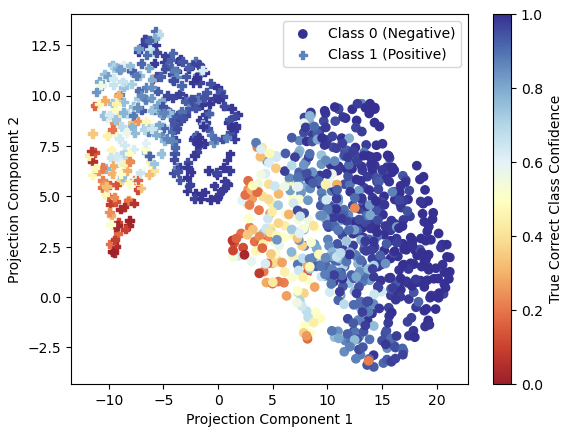}}

  \caption{Comparison of model performance trends within various embedding spaces. Only the Anchor Points Map results in clean localization of model performance where a held-out model's performance on each point correlates with its performance on neighboring points. This allows fine-grained comparison of how different models perform on the same dataset distribution.} 
  \label{fig:embedding-comparison}
  
\end{figure*}

\begin{figure*}[]
  \centering
  
  \subfigure[QQP Anchor Point Map Colored with Estimated text-curie-001 (Prompt 1) predictions. The estimates achieve a large MAE of 0.30 and poor agreement of 54\%. \label{fig:curie-est}]{\includegraphics[scale=0.32]{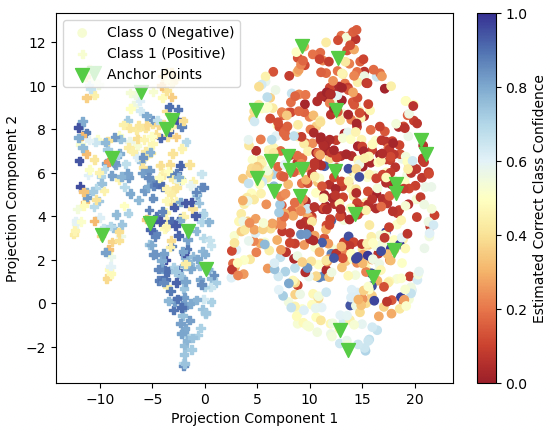}} \quad
  \subfigure[QQP Anchor Point Map Colored with True text-curie-001 (Prompt 1) predictions.\label{fig:curie-true}]{\includegraphics[scale=0.32]{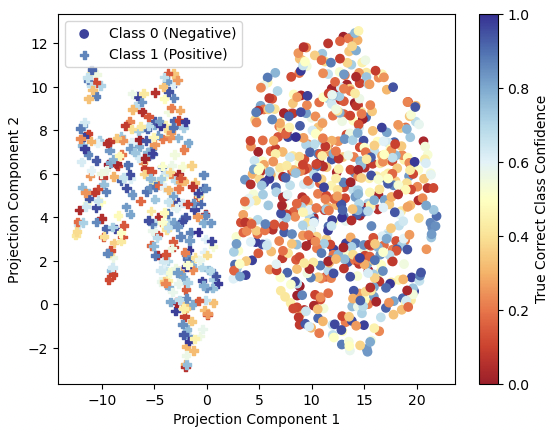}}

  \caption{Anchor Point Failure Case: The map is computed using the predictions of 60 randomly-selected source models on QQP and used to estimate the predictions of text-curie-001 with Prompt 1 (\ref{fig:curie-est}). The estimates achieve a large MAE and near random agreement. Upon inspecting the true predictions (\ref{fig:curie-true}), we observe that the behavior of text-curie-001 is not localized in the Anchor Point Map. Model performance appears sporadic. This suggests that text-curie-001 does not follow the same predictive correlations as the source models, preventing effective estimation of the model's predictions.} 
  \label{fig:curie-failure}
  
\end{figure*}

\begin{figure*}
\centering
\includegraphics[scale=0.5]{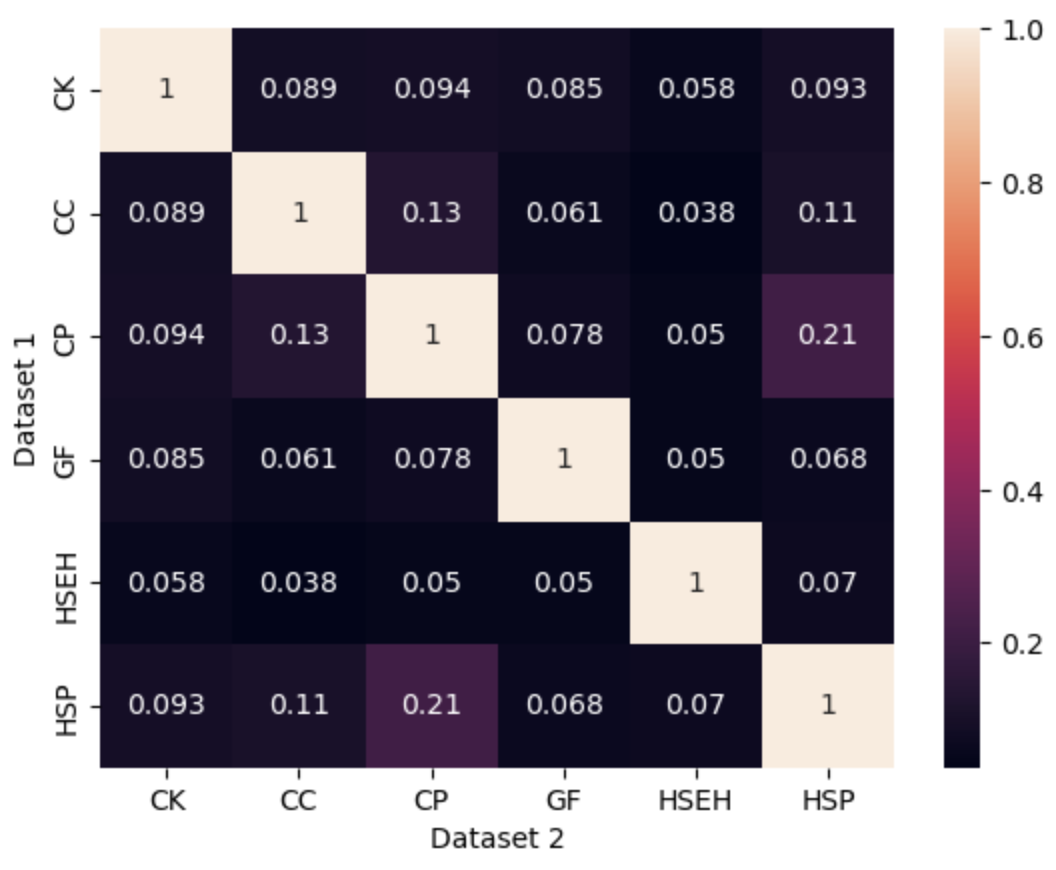}

\caption{\label{fig:vocab-overlap} Vocabulary overlap of our MMLU datasets. Each cell indicates the portion of words in dataset 1 that are also in dataset 2. Labels: CK = clinical knowledge, CC = college chemistry, CP = college physics, GF = global facts, HSEH = high school European history, HSP = high school physics.}

\end{figure*}

\begin{table*}[t]
\addtolength{\tabcolsep}{-0.1em}
\centering
\footnotesize
\begin{tabular}{l@{\ }c|cccccc|cc}
\toprule
&  & Random & Random &
Pretrained & Pretrained  & Fine-Tuned & Fine-Tuned & AP & AP \\
& Mean &  & Weighted & & Weighted & & Weighted &  Weighted & Predictor  \\
\midrule 
Datasets & Family &  Exact & Corr &  Exact & Corr & Exact & Corr & Corr & Exact \\
\midrule 
SST-2  & BERT & 0.485 & 0.607  & 0.512  & 0.637 & 0.594  & \textbf{0.738}  & \underline{0.696}  & 0.593 \\
QQP  & BERT & 0.459 & 0.631 & 0.670 & 0.613 & 0.572 & \textbf{0.756} & \underline{0.691} & 0.555 \\
RTE  & BERT & 0.478 & 0.406 & 0.316 & -0.001 & 0.486 & 0.335 & \underline{0.591} & \textbf{0.597} \\
QNLI & BERT & 0.346 & 0.523 & -0.252 & 0.613 & -0.430 & 0.510 & \textbf{0.665} & \underline{0.571} \\
MRPC  & BERT & 0.730 & 0.741 & 0.737 & 0.749 & 0.692 & \textbf{0.786} & \underline{0.778} & 0.721 \\
MNLI & BERT & 0.684 & 0.748 & 0.646 & 0.653 & 0.676 & \textbf{0.735} & \underline{0.733} & 0.720 \\
\midrule 
\textbf{BERT Avg.} &  & 0.530 & 0.609 & 0.438 & 0.543 & 0.432 & \underline{0.643} & \textbf{0.699} & 0.626 \\
\midrule 
SST-2  & GPT & 0.440 & 0.278  & 0.548  & 0.215 & \textbf{0.651}  & 0.531  & 0.441  & \underline{0.632} \\
QQP  & GPT & 0.561 & 0.591 & -0.557 & -0.537 & 0.720 & 0.752 & \underline{0.777} & \textbf{0.825} \\
RTE  & GPT & 0.188 & 0.138 & -0.119 & -0.251 & 0.133 & 0.082 & \underline{0.378} & \textbf{0.452} \\
QNLI & GPT & 0.101 & 0.076 & 0.209 & 0.197 & 0.185 & \underline{0.130} & \textbf{0.158} & 0.126 \\
MRPC  & GPT & 0.475 & 0.460 & 0.155 & 0.321 & 0.360 & 0.360  & \underline{0.516} & \textbf{0.681} \\
MNLI & GPT & 0.033 & 0.059 & -0.517 & -0.382 & 0.181 & 0.099 & \underline{0.201} & \textbf{0.494} \\
\midrule 

\textbf{GPT Avg.} & GPT & 0.300 & 0.267 & -0.470 & -0.730 & 0.371 & 0.325 & \underline{0.410} & \textbf{0.535} \\

\midrule 
SST-2  & IGPT & 0.655 & 0.722  & 0.726  & 0.728 & 0.606  & \underline{0.760}  & \textbf{0.814}  & 0.719 \\
QQP  & IGPT & 0.734 & 0.801 & 0.264 & 0.276 & 0.824 & \textbf{0.893} & \underline{0.884} & 0.845 \\
RTE  & IGPT & 0.389 & 0.322 & 0.220 & 0.03 & 0.391 & 0.323 & \textbf{0.671} & \underline{0.629} \\
QNLI & IGPT & 0.404 & 0.400 & 0.512 & 0.469 & 0.449 & 0.296 & \underline{0.597} & \textbf{0.690} \\
MRPC  & IGPT & 0.727 & 0.680 & 0.520 & 0.651 & 0.651 & 0.712 & \textbf{0.776} & \underline{0.738} \\
MNLI & IGPT & 0.265 & 0.305 & -0.074 & -0.088 & \textbf{0.460} & 0.420 & \underline{0.434} & 0.317 \\
\midrule
\textbf{IGPT Avg.} &  & 0.529 & 0.538 & 0.361 & 0.344 & 0.563 & 0.566 & \textbf{0.696} & \underline{0.656} \\

\midrule 
SST-2  & OAI & 0.666 & 0.697  & 0.774  & \textbf{0.871} & 0.761  & \underline{0.833}  & 0.734  & 0.661 \\
QQP  & OAI & 0.610 & 0.651& -0.090 & -0.080 & 0.676 & \underline{0.722} & \textbf{0.747} & 0.637 \\
RTE  & OAI & 0.362 & 0.303 & 0.206 & 0.066 & \underline{0.381} & 0.294 & -0.378 & \textbf{0.411} \\
QNLI & OAI & 0.451 &\underline{0.453} & 0.381 & 0.378 & 0.286 & -0.03 & \textbf{0.497} & 0.309 \\
MRPC  & OAI & 0.633 & 0.661 & 0.533 & 0.563 & 0.550 & 0.673 & \textbf{0.714} & \underline{0.705} \\
MNLI & OAI & 0.300 & 0.278 & -0.015 & -0.181 & 0.559 & \textbf{0.490} & 0.350 & \underline{0.486} \\
\midrule
\textbf{OAI Avg.} & OAI & 0.503 & 0.507 & 0.298 & 0.270 & \underline{0.536} & 0.497 & \textbf{0.570} & 0.535 \\

\midrule 
SST-2 & ALL & 0.685 & 0.705  & 0.734  & 0.725 & 0.730  & \textbf{0.787}  & \underline{0.757}  & 0.727 \\
QQP & ALL & 0.669 & 0.678 & 0.189 & 0.233 & \underline{0.766} & \textbf{0.770} & 0.756 & 0.701 \\
RTE & ALL & 0.366 & 0.308 & 0.143 & -0.052 & 0.354 & 0.275 & \textbf{0.483} & \underline{0.462} \\
QNLI & ALL  & 0.321 & \underline{0.331} & 0.192 & 0.294 & 0.127 & 0.144 & \textbf{0.439} & 0.303\\
MRPC & ALL & 0.687 & 0.679 & 0.528 & 0.604 & 0.641 & 0.681 & \textbf{0.726} & \underline{0.716} \\
MNLI& ALL & 0.438 & 0.433 & 0.177 & 0.166 & 0.523 & 0.453 & \textbf{0.544} & \underline{0.517} \\
\midrule
\textbf{Avg.} & ALL & 0.528 & 0.522 & 0.327 & 0.328 & 0.523 & 0.518 & \textbf{0.612} & \underline{0.571} \\

\bottomrule
\end{tabular}
\caption{Area Under the (Kendall's $\tau$) Correlation Curve from 1 to 30 points for ranking language models at 6 GLUE tasks from each model family. We randomly select 5 source models each for the InstructGPT and OpenAI Families and 10 each for the BERT, GPT, and ALL families. We then rank the remaining models within each family and average over 100 randomized runs. AP Weighted and AP Predictor prove to be the most effective at accurately ranking models in this small-data regime. In each row, the best score is \textbf{bolded} and second best score is \underline{underlined}.} 
\label{tab:AUCC_table_families}
\end{table*}

\section{Anchor Points for Ranking Massive Multitask Language Understanding}
\label{apppendix:mmlu}

\begin{table}[H]
\centering
\footnotesize
\begin{tabular}{l@{\ }c}
\toprule
Model \\
\midrule 
1. huggyllama/llama-7b \\
2. huggyllama/llama-13b\\
3. mosaicml/mpt-7b   \\
4. tiiuae/falcon-7b   \\
5. facebook/opt-350m   \\
6. facebook/opt-125m    \\
7. facebook/opt-6.7B  \\
8. mistralai/Mistral-7B-Instruct-v0.1    \\
9. Eleuther/pythia-12b      \\
10. openlm-research/open\_llama\_7b    \\
11. meta-llama/Llama-2-7b   \\
12. openlm-research/open\_llama\_3b    \\
13. openlm-research/open\_llama\_7b \\
14. openlm-research/open\_llama\_13b   \\

\bottomrule
\end{tabular}
\caption{LLMs used for MMLU experiments. Models can be accessed at https://huggingface.co/models.}
\label{tab:mmlu-models}
\end{table}

We perform a small scale experiment to evaluate anchor points at ranking models on 6 Massive Multitask Language Understanding \cite{mmlu} datasets: clinical knowledge, college chemistry, college physics, global facts, high school European history, and high school physics. Each dataset contains multiple choice questions requiring extensive real-world knowledge.  Table \ref{tab:AUCC_table_MMLU} shows results of the Anchor Point Predictor, Anchor Point Weighted, and baselines at ranking 7 target models using 7 source models with 1-30 anchor points. Due to the complexity of these tasks, we substitute the fine-tuned BERT embedding baselines with BAAI/bge-large-en-v1.5, a state-of-the-art embedding model. 

We find that Anchor Points Weighted is the strongest performer overall, outperforming random selection on 5 of the 6 datasets with an average improvement of 0.07 AUCC. However, the Anchor Points Predictor outperforms random on only 3 of 6 datasets. Notably, both methods are far below random on the Global Facts dataset. This is perhaps due to the randomness and specificity of the Global Facts questions: many have answers that are likely to be rare in pre-training data, such as 'What is the percentage of children aged 13-15 in the United States who reported being bullied at least once in the past couple of months as of 2015?" The correlations learned between these questions (using only 7 source models) are likely to be weak or spurious. Increasing the source model set size would improve the probability of learning generalizable predictive correlations.

Interestingly, the improved performance of Pretrained Weighted in MMLU relative to GLUE suggests that semantic similarity in questions can be an effective proxy of model performance similarity when questions emphasize intensive knowledge rather than simple language understanding. However, the inconsistency of these approaches still warrants a more reliable technique such as Anchor Points Weighted.

\section{Hyperparameters, Compute, and Packages}
\label{appendix:hyperparameters}

For each GLUE task, we finetune all BERT-family with a batch size of 32, learning rate of 2e-5, and weight decay of 0.01 for 3 epochs. We did not perform extensive hyperparameter tuning; thus, model performances do not necessarily represent their ideal performance. We chose to not perform hyperparameter tuning in order to assess whether anchor points could reliably evaluate models with a wide range of performances. This training process, as well as the process of training code development and model inference, took approximately 120 GPU hours with a single A100.

We used Sentence Transformers 2.2.2, Scipy 1.10.1, transformers 4.25.1, tokenizers 0.11.4, numpy 1.23.5, scikit-learn 1.2.2, kmedoids 0.4.3, and the Eleuther Evaluation Harness (https://github.com/EleutherAI/lm-evaluation-harness).

\section{Prompts}
\label{appendix:prompts}
We enumerate the zero-shot prompts used for each dataset below. All prompts are randomly selected from Prompt Source \cite{promptsource}

\paragraph{QQP}

\begin{enumerate}
    \item Can an answer to "\{\{question1\}\}" also be used to answer "\{\{question2\}\}"?
    \item I received the questions "\{\{question1\}\}" and "\{\{question2\}\}". Are they
      duplicates? 
    \item Are the questions "\{\{question1\}\}" and "\{\{question2\}\}" asking the same thing?
\end{enumerate}

\paragraph{SST-2}

\begin{enumerate}
    \item Does the following sentence have a \{\{"positive"\}\} or \{\{"negative"\}\} sentiment? \{\{sentence\}\}
    \item Someone just said to me "\{\{sentence\}\}". Do you think they are \{\{"sad"\}\} or \{\{"happy"\}\}?'
    \item I'm reading a review that says "\{\{sentence\}\}".
      Do you think the review is \{\{"positive"\}\} or \{\{"negative"\}\}?
\end{enumerate}

\paragraph{RTE}

\begin{enumerate}
    \item Does the claim "\{\{sentence2\}\}" follow from the fact that "\{\{sentence1\}\}"?
      Please answer either \{\{"yes"\}\} or \{\{"no"\}\}
    \item Is the relationship from the first to the second sentence "\{\{"entailment"\}\}"
      or "\{\{"not entailment"\}\}"?
    \item Does "\{\{sentence1\}\}" imply that "\{\{sentence2\}\}"? Please answer either
      \{\{"yes"\}\} or \{\{"no"\}\}
\end{enumerate}

\paragraph{QNLI}

\begin{enumerate}
    \item Can you answer the question "\{\{question\}\}" based only on the following:
      \{\{sentence\}\}
    \item \{\{sentence\}\}
      Does that sentence have all you need to answer the question "\{\{question\}\}"?
    \item Does knowing that "\{\{sentence\}\}" imply that I know the answer to "\{\{question\}\}
\end{enumerate}

\paragraph{MNLI}

\begin{enumerate}
    \item Suppose it's true that \{\{premise\}\} Then, is "\{\{hypothesis\}\}" \{\{"always"\}\},
      \{\{"sometimes"\}\}, or \{\{"never"\}\} true?
    \item Question: \{\{hypothesis\}\} True, False, or Neither?
    \item \{\{premise\}\} Using only the above description and what you know about the
      world, "\{\{hypothesis\}\}" is definitely correct, incorrect, or inconclusive?
\end{enumerate}

\paragraph{MRPC}

\begin{enumerate}
    \item \{\{sentence1\}\} paraphrase (that is, mean the same thing as) this sentence?
        \{\{sentence2\}\}
    \item Can I replace the sentence \{\{sentence1\}\} with the sentence \{\{sentence2\}\} and have it mean the same thing?
    \item Are the following two sentences "\{\{"equivalent"\}\}" or "\{\{"not equivalent"\}\}"? \{\{sentence1\}\} \{\{sentence2\}\}
\end{enumerate}

\label{sec:appendix}

\begin{algorithm*}
\caption{Anchor Points Predictor Fit}\label{alg:fit}
\begin{algorithmic}

\State $K \gets$ number of anchor points
\State $P \gets N x D x Q$ tensor of $N$ source models' output probabilities on $D$ points over $Q$ classes
\State $P \gets logit(P)$ 
\Comment{take logit transform of predictions}

\State $M \gets$ empty $(D - K) \times K \times Q$  array \Comment{To store slopes}
\State $B \gets$ empty $(D - K) \times K \times Q$ array \Comment{To store biases}
\State $R \gets$ empty $(D - K) \times K \times Q$ array \Comment{To store residuals}
\State $N \gets$ empty $(D - K) \times K \times Q$ array \Comment{Zero-initialized tensor to indicate nearest anchors}

\State $C \gets$ sum([corrcoef($P[:,:,i]$ for $i$ in range($Q$)]) / $Q$
\Comment{$D \times D$  averaged correlation matrix}

\State $AP \gets $ K-MEDOIDS(1 - C, K) 
\Comment{length K array of anchor 
point indices}

\State $T \gets \{0...D - 1\}$ \textbackslash $AP$ \Comment{length D - K array of test point indices}

\For{q in range(Q)}
\For{i, test $\in$ enumerated $T$}
\For{j,anchor $\in$ enumerated $AP$}
\State $x \gets$ P[:,j,q]
\State $y \gets$ P[:,i,q]
\State m, b, r $\gets$ LinearRegression(x, y) 
\Comment{Slope, bias, and residual of trend line}
\State M[i,j,q] $\gets$ m
\State B[i,j,q] $\gets$ b
\State R[i,j,q] $\gets$ r

\EndFor
\EndFor

\State nearest $\gets$ argmin(R[:,:,q], axis = 1) \Comment{nearest anchor indices to each point}
\State N[arange(D-K),nearest,q] $\gets$ 1 \Comment{populate indicator array}
\EndFor

\Return AP, N, T, M, B

\end{algorithmic}
\end{algorithm*}

\begin{algorithm*}
\caption{Anchor Points Predictor Predict}\label{alg:predict}
\begin{algorithmic}

\Require AP, N, T, M, B \Comment{Returned by Anchor Points Fit}
\State P $\gets$ length K array of a target model's gold label predictions on the K anchor points
\State $P \gets$ logit(P)
\Comment{take logit transform of predictions}
\Comment{To store estimated target model predictions}
\State preds $\gets$ (M * P[newaxis,:,:] + B) \Comment{Prediction step}
\State preds $\gets$ sum(preds * N, axis = 1) \Comment{Prune predictions from non-nearest anchors}
\State Y $\gets$ expit(preds) \Comment{Inverse of the logit transform}

\Return Y 

\end{algorithmic}
\end{algorithm*}

\begin{figure*}[t]
  \centering
  \subfigure[MNLI] {\includegraphics[scale=0.35]{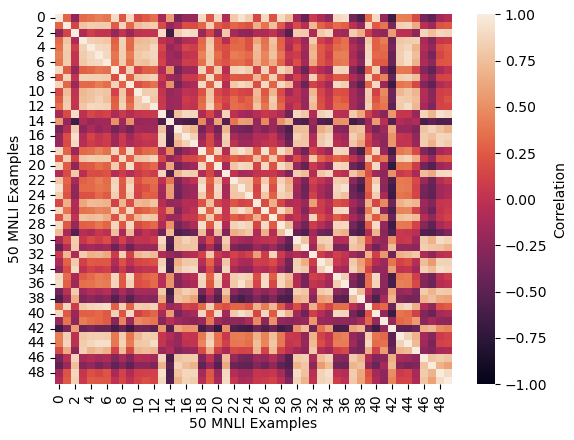}} \quad
  \subfigure[MRPC]{\includegraphics[scale=0.35]{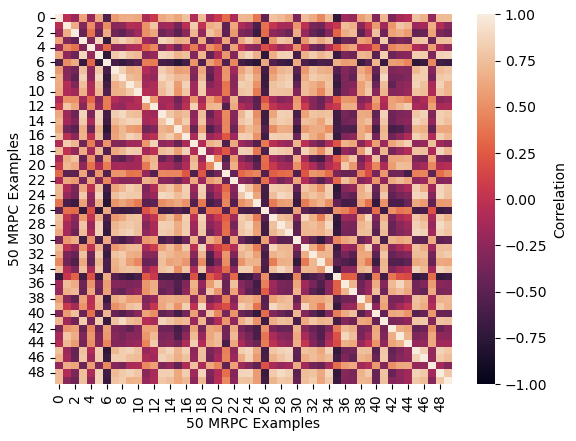}}
  \subfigure[QNLI]{\includegraphics[scale=0.35]{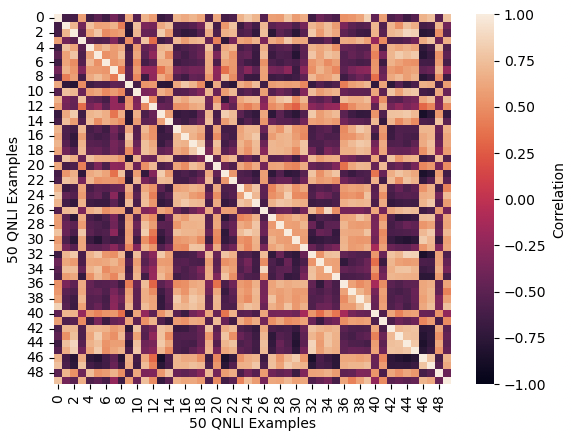}} \quad
  \subfigure[QQP]{\includegraphics[scale=0.35]{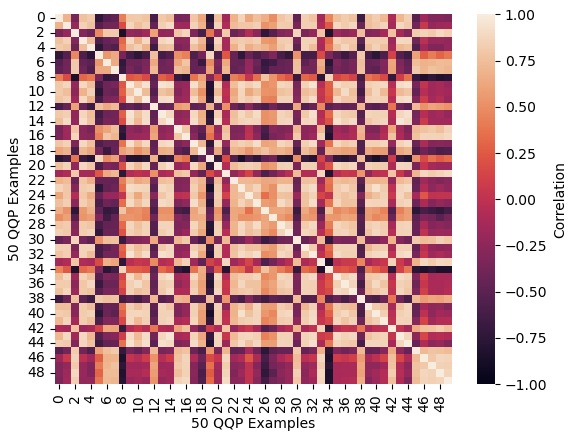}}
  \subfigure[RTE]{\includegraphics[scale=0.35]{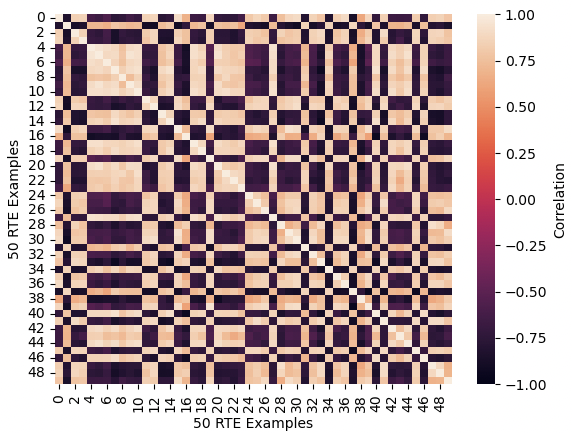}} \quad
  \subfigure[SST-2]{\includegraphics[scale=0.35]{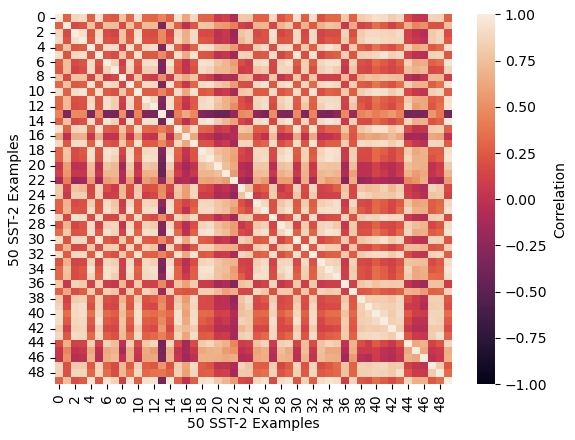}}
  \caption{Correlation Matrices of Correct Class Confidence Predictions from 87 models across all model families. Each plot shows the correlations between 50 examples sampled from the validation Sets of GLUE Tasks.}
  \label{fig:corr-mats-gpt}
  
\end{figure*}

\begin{table*}[t]
\centering
\footnotesize
\begin{tabular}{l@{\ }c|cccccc}
\toprule
Model & Parameter Count & MNLI & SST-2 & QQP & RTE & MRPC & QNLI  \\
\midrule 
1. Bio-ClinicalBERT  & 110M & 0.502  & 0.875  & 0.815 & 0.472  & 0.384 & 0.595   \\
2. albert-base-v2  & 11M & 0.325  & 0.927  & 0.797 & 0.472  & 0.522 &0.505   \\
3. bart-base  & 139M & 0.834  & 0.954  & 0.841 & 0.747  & 0.855 & 0.626   \\
4. bart-large  & 400M & 0.891  & 0.915  & 0.639 & 0.527  & 0.892 & 0.505   \\
5. bert-base-cased  & 110M & 0.444  & 0.917  & 0.639 & 0.519  & 0.504 & 0.505   \\
6. bert-base-multilingual-cased  & 110M & 0.478  & 0.890  & 0.828 & 0.472  & 0.392 & 0.505   \\
7. bert-base-uncased  & 110M & 0.468  & 0.917  & 0.639 & 0.476  & 0.318 & 0.630   \\
8. bert-large-cased  & 336M & 0.319  & 0.509  & 0.639 & 0.487  & 0.627 & 0.505   \\
9. bert-large-uncased  & 336M & 0.611  & 0.916  & 0.639 & 0.472  & 0.387 & 0.505   \\
10. bert-mini  & 11M & 0.479  & 0.845  & 0.819 & 0.476  & 0.664 & 620   \\
11. bert-tiny  & 3M & 0.428  & 0.803  & 0.784 & 0.530  & 0.479 & 0.639   \\
12. biobert-v1.1  & 110M & 0.539  & 0.894  & 0.825 & 0.50  & 0.540 & 0.617   \\
13. deberta-base  & 100M & 0.876  & 0.940  & 0.639 & 0.631 & 0.884 & 0.505   \\
14. deberta-large  & 350M & 0.344  & 0.950  & 0.639 & 0.527  & 0.904 & 0.505   \\
15. deberta-v3-base  & 304M & 0.320  & 0.950  & 0.850 & 0.837  & 0.880 & 0.663   \\
16. deberta-v3-xsmall & 22M & 0.352  & 0.924  & 0.850 & 0.678  & 0.865 & 0.660   \\
17. distilbert-base-cased  & 67M & 0.820  & 0.90  & 0.831 & 0.570  & 0.818 & 0.618   \\
18. distilbert-base-uncased  & 66M & 0.820  & 0.896  & 0.840 & 0.570  & 0.830 & 0.613  \\
19. electra-base-discriminator  & 102M & 0.863  & 0.943  & 0.830 & 0.761  & 0.892 & 0.645   \\
20. legal-bert-small-uncased  & 24M & 0.415  & 0857  & 0.831 & 0.472  & 0.450 & 0.597   \\
21. roberta-base  & 125M & 0.873  & 0.938  & 0.834& 0.736  & 0.870 & 0.505   \\
22. scibert-scivocab-uncased  & 110M & 0.528  & 0.891  & 0.830 & 0.550  & 0.436 & 0.599   \\
23. sentence-bert  & 110M & 0.413  & 0.919  & 0.8225 & 0.545  & 0.321 & 0.505   \\
24. sentiment-roberta-large-english  & 335M & 0.890  & 0.951  & 0.693 & 0.580  & 0.840 & 0.505   \\
25. twitter-roberta-base  & 125M & 0.835  & 0.930  & 0.833 & 0.588  & 0.855 & 0.590   \\
26. xlm-roberta-base  & 279M & 0.835  & 0.916  & 0.633 & 0.527  & 0.860 & 0.505   \\
27. xlm-roberta-large  & 355M & 0.353  & 0.509  &0.639 & 0.602  & 0.884 & 0.505   \\

\bottomrule
\end{tabular}
\caption{BERT-Family Model Accuracies on Six GLUE Tasks. Hyperparameters are in Appendix \ref{appendix:hyperparameters}. Models can be accessed at https://huggingface.co/models.}
\label{tab:accs1}

\end{table*}

\begin{table*}[t]
\centering
\footnotesize
\begin{tabular}{l@{\ }c}
\toprule
Model & Parameter Count \\
\midrule 
1. Cerebras-GPT-1.3B  & 1.3B \\
2. Cerebras-GPT-111M  &  111M\\
3. Cerebras-GPT-256M   &  256M \\
4. bloom-1b7  & 1.72B    \\
5. gpt-neo-1.3B  & 1.3B   \\
6. gpt-neo-125m  & 125M   \\
7. gpt2-large  & 774M  \\
8. gpt2-medium  & 355M   \\
9. gpt2  & 137M     \\
10. openai-gpt  & 120M   \\
\midrule 
11. RedPajama-INCITE-Instruct-7B-v0.1 & 7B \\
12. falcon-7b-instruct  & 7B \\
13. mpt-7b-instruct  & 7B \\
14. mt0-xl  & 3.7B \\
15. bloomz-3b  & 3B \\
\midrule 
16. text-ada-001 & Not publicly known  \\
17. text-babbage-001 & Not publicly known   \\
18. text-curie-001& Not publicly known    \\
19. text-davinci-002 & Not publicly known   \\
20. text-davinci-003 & Not publicly known   \\

\bottomrule
\end{tabular}
\caption{GPT, InstructGPT, and OpenAI Family Model Zero-Shot Models and Parameter Counts (Used for GLUE experiments). OpenAI Model parameter counts are unknown. Models can be accessed at https://huggingface.co/models and https://platform.openai.com/docs/models.}
\label{tab:accs2}
\end{table*}

\end{document}